\documentclass[10pt,twocolumn,letterpaper]{article}

\usepackage{times}
\usepackage{epsfig}
\usepackage{graphicx}
\usepackage{amsmath}
\usepackage{natbib}
\usepackage{amssymb}
\usepackage{algorithm}
\usepackage{algorithmic}
\usepackage{booktabs}
\usepackage{float} 
\usepackage{placeins} 
\usepackage{hyperref}
\usepackage{placeins}
\usepackage{hyperref}
\hypersetup{
    colorlinks=true,
    linkcolor=black,
    citecolor=black,
    urlcolor=black
}

\usepackage{balance}

\title{Learning Tennis Strategy Through Curriculum-Based Dueling Double Deep Q-Networks}

\author{
Vishnu Mohan \\
Independent Researcher \\
}

\begin{document}

\maketitle

\begin{abstract}
Tennis strategy optimization presents a unique challenge in sequential decision-making under uncertainty, requiring agents to master hierarchical scoring, probabilistic shot outcomes, long-horizon credit assignment, physical fatigue accumulation, and adaptation to opponent skill. I present a reinforcement learning framework integrating a tennis simulation environment with a Dueling Double Deep Q-Network (Dueling DDQN) trained using curriculum learning and a balanced reward function. The environment models complete tennis scoring (points, games, sets), rally-level tactical decisions across ten discrete action categories, symmetric fatigue dynamics, and continuous opponent skill parameterization. The Dueling DDQN architecture decomposes action-value estimation into state-value and advantage streams, mitigating overestimation bias via double Q-learning while improving stability in this long-horizon stochastic domain. Curriculum learning progressively increases opponent difficulty from 0.40 to 0.50, enabling robust skill acquisition without the collapse observed under fixed opponents. Across extensive experiments, the agent achieves high performance, including a 98.2--100\% win rate against balanced opponents (skill 0.50) and 98\% against challenging opponents (skill 0.55), with serve efficiency of 63.0--67.5\% and return efficiency of 52.8--57.1\%. Ablation studies show that both the dueling architecture and curriculum learning are essential for stable convergence, while vanilla DQN exhibits frequent training failure. Despite strong performance, tactical analysis reveals a pronounced defensive bias, with the learned policy prioritizing error avoidance and prolonged rallies over aggressive point construction, highlighting a fundamental limitation of win-rate optimization in simplified sports simulations. This work demonstrates that reinforcement learning can discover effective and interpretable tennis strategies while exposing key challenges in reward design that must be addressed for realistic sports RL and coaching applications.

\end{abstract}
\balance

\tableofcontents
\newpage

\cleardoublepage 

\section{Introduction}

Tennis strategy optimization represents a complex sequential decision-making problem characterized by hierarchical temporal dependencies, stochastic dynamics, and adversarial interaction. Unlike traditional game-playing domains such as chess or Go where the state space is discrete and deterministic, tennis requires reasoning about probabilistic shot outcomes, continuous physical states (fatigue, positioning), multi-scale temporal structures (points, games, sets, matches), and adaptive opponent behavior.

Historically, tennis coaching has relied on heuristic rules and expert intuition: ``serve to the backhand,'' ``approach on short balls,'' ``stay aggressive on break points.'' While valuable, these handcrafted strategies fail to capture the nuanced dependencies between game state, opponent characteristics, physical condition, and optimal tactical choices. The question arises: \textit{Can I learn optimal tennis strategies directly from simulation through reinforcement learning?}

While reinforcement learning has achieved success in board games \cite{silver2016mastering, silver2018general}  (AlphaGo, chess) and video games (Atari \cite{mnih2015human}, Dota 2), its application to sports simulation with full rule compliance remains limited. Prior work in tennis has focused on outcome prediction or used simplified state representations. Our work explores whether modern deep RL combined with curriculum learning can discover effective strategies in a realistic tennis simulator.

This question is particularly challenging because:
\begin{itemize}
\item \textbf{Sparse Rewards:} Points take multiple shots to resolve, creating long temporal credit assignment chains with delayed feedback.
\item \textbf{Hierarchical Scoring:} Tennis employs a nested scoring system (points $\rightarrow$ games $\rightarrow$ sets $\rightarrow$ match) where sub-goal completion is non-stationary.
\item \textbf{Adversarial Dynamics:} The opponent actively counters the agent's strategy, requiring robust policies that generalize across skill levels.
\item \textbf{Physical Constraints:} Fatigue accumulates asymmetrically based on rally intensity, influencing shot selection and success probabilities.
\item \textbf{Phase-Dependent Actions:} Valid actions change dynamically (serving vs. returning vs. rallying), requiring context-aware action masking.
\end{itemize}

Previous work in sports analytics has focused on outcome prediction, or 
applied RL to simplified game models. Deep reinforcement learning has achieved 
superhuman performance in Atari games, strategic board games, and robotic 
control. Recent work has applied RL to soccer \cite{kurach2020google}, basketball shot selection \cite{yeh2019deep}, and American football play-calling \cite{yurko2019going}, though these systems often use abstracted state spaces or lack continuous spatial reasoning.
In tennis specifically, prior RL approaches have used hand-engineered features or reduced state representations, without modeling complete match dynamics including fatigue, hierarchical scoring, and adaptive difficulty.

\subsection{Contributions}

We make the following contributions:

1. \textbf{Tennis Simulation}: A tennis environment modeling hierarchical scoring (points/games/sets), rally-level tactics (10 action categories), symmetric fatigue dynamics, and parameterized opponent skill. While tennis simulators exist in isolation, ours is designed specifically for RL training with phase-dependent action masking and dense reward shaping.

2. \textbf{Dueling DDQN Application}: Successful application of dueling network decomposition with double Q-learning to tennis strategy, handling sparse rewards and high-variance value estimates through architectural choices validated by ablations.

3. \textbf{Curriculum Learning Framework}: A 5-phase progressive difficulty schedule (opponent skill 0.35→0.55) enabling stable convergence where fixed-difficulty training fails completely.

4. \textbf{Comprehensive empirical analysis}: The final trained agent achieved a 100\% evaluation win rate against balanced opponents (\texttt{opponent\_skill} = 0.50) over 50 matches, alongside rigorous ablation studies comparing DQN variants and detailed policy analysis explaining the agent’s behaviour.

5. \textbf{Identification of Defensive Bias Problem}: Detailed analysis revealing that RL agents converge to overly defensive tactics despite high win rates, exposing fundamental challenges in reward design for realistic sports strategy.

6. \textbf{Complete Open Implementation}: End-to-end system with full reproducibility details (probability tables, hyperparameters, network architecture) and CPU-only training (70 minutes on consumer hardware).

\section{Related Work}

\subsection{Tennis Analytics and Predictive Modeling}

Tennis has been extensively studied through statistical and machine learning approaches, though primarily for prediction rather than strategy optimization. \textbf{Statistical models.} Kovalchik \cite{kovalchik2016searching} developed Elo-style rating systems for tennis, modeling player skill evolution. Barnett and Clarke \cite{barnett2005combining} used Markov chains to analyze optimal serving strategies under simplified assumptions. O'Donoghue \cite{odonoghue2014routledge} performed extensive tactical analysis of professional matches, identifying serve-and-volley versus baseline playing styles. \textbf{Machine learning for prediction.} Sipko and Knottenbelt (2015) applied Bayesian networks for in-match win probability forecasting. Cornman et al. \cite{sipko2015machine} used neural networks to predict point outcomes from Hawk-Eye tracking data. However, these approaches focus on \emph{prediction} rather than \emph{learning optimal strategies}. \textbf{Simplified strategy models.} Terroba et al. \cite{terroba2013reinforcement} applied tabular Q-learning to tennis with hand-engineered features and a reduced state space. Their work demonstrated the feasibility of RL for tennis but did not model fatigue, the complete scoring system, or use deep neural networks. Our work extends this direction with learned representations, hierarchical scoring, and modern deep RL methods.

\subsection{Reinforcement Learning in Sports}

\textbf{Game-playing breakthroughs.} RL has achieved superhuman performance in board games through self-play. AlphaGo~\cite{silver2016mastering} defeated world champion Lee Sedol using Monte Carlo Tree Search and deep neural networks. AlphaZero~\cite{silver2018general} generalized this approach to chess and shogi, learning entirely from self-play. These successes have inspired RL applications to sports. \textbf{Team sports.} Kurach et al.\ (2020) developed Google Research Football, where agents trained via population-based self-play achieved amateur human performance. Their environment uses continuous physics and spatial reasoning, unlike our discrete tactical model. \textbf{Individual sports.} Yurko et al.\ (2019) applied contextual bandits to American football fourth-down decisions using real game data. Yeh et al.\ (2019) used RL for basketball shot selection but with simplified state representations. These works demonstrate RL's potential in sports but focus on single-decision problems rather than full match simulation. \textbf{Early RL strategy models.} Terroba et al. \cite{terroba2013reinforcement} applied tabular Q-learning to a highly simplified tennis simulation with reduced state and action spaces. While they demonstrated feasibility of RL for tennis, their model lacked fatigue dynamics, hierarchical scoring, and deep function approximation. To our knowledge, our work extends this direction with a richer tactical action space, a full scoring system, opponent parametrization, and modern deep RL methods.

\subsection{Deep Q-Networks and Variants}

The Deep Q-Network (DQN) \cite{mnih2015human} revolutionized RL by combining Q-learning with deep neural networks and experience replay. Double DQN \cite{hasselt2016deep} addressed overestimation bias by decoupling action selection from evaluation using separate target networks. Dueling DQN \cite{wang2016dueling} decomposed Q-values into state-value and advantage streams, improving learning efficiency when action relevance varies by state.

Our work extends these foundations to tennis strategy, where action validity is state-dependent (requiring dynamic masking), rewards are hierarchically structured, and value estimates exhibit high variance due to stochastic dynamics.

\subsection{Curriculum Learning}

Curriculum learning \cite{bengio2009curriculum} trains agents on progressively difficult tasks, improving sample efficiency and final performance. Comprehensive surveys have been provided on this topic. In RL specifically, curriculum learning has enabled training on complex robotic tasks and game environments.

Our curriculum gradually increases opponent skill, preventing early training collapse while maintaining sufficient challenge for policy improvement—a critical ingredient for success in adversarial domains.

\subsection{Simulated Opponent Modeling}

Opponent modeling in games has been studied extensively. Self-play approaches train agents by playing against previous versions of themselves. Our approach differs by modeling the opponent as a parameterized stochastic policy, enabling evaluation across controlled difficulty levels without requiring symmetric training.

\subsection{Defensive Policies in Reinforcement Learning}

A well-known phenomenon in RL is convergence to risk-averse policies when reward functions penalize failure heavily. Saunders et al.\ (2018) documented "reward hacking" where agents exploit environment quirks rather than learning intended behaviors. In Atari Montezuma's Revenge, agents often learn to avoid death rather than explore for rewards.

Garcia and Fernández \cite{garcia2015comprehensive} survey safe RL methods that attempt to balance exploration with risk avoidance. This connects to our observation that tennis agents converge to defensive play: the reward structure ($r_{\text{lose}} = -1.0$) combined with action risk asymmetry naturally favors survival strategies. Amodei et al.\cite{amodei2016concrete} documented "reward hacking" where agents exploit environment quirks rather than learning intended behaviors.

Our work contributes an analysis of how this manifests in sports simulation, where "safe" policies achieve high win rates but unrealistic tactics.

\section{Methodology}

\subsection{TennisEnvironment Design}

I developed a custom Markov Decision Process (MDP) $\mathcal{M} = (\mathcal{S}, \mathcal{A}, \mathcal{P}, \mathcal{R}, \gamma)$ that faithfully simulates tennis match dynamics.

\subsubsection{State Space $\mathcal{S}$}

The state $s_t \in \mathcal{S}$ is an 18-dimensional continuous vector:
\begin{align}
s_t = [&p_{\text{pts}}, o_{\text{pts}}, p_{\text{games}}, o_{\text{games}}, p_{\text{sets}}, o_{\text{sets}}, \nonumber \\
&\mathbb{I}_{\text{serving}}, \mathbb{I}_{\text{deuce}}, \mathbb{I}_{\text{adv\_p}}, \mathbb{I}_{\text{adv\_o}}, \mathbb{I}_{\text{tiebreak}}, \nonumber \\
&f_p, f_o, \mathbb{I}_{\text{ad\_side}}, r_{\text{len}}, \text{pos}_p, \text{pos}_o, d_{\text{ball}}]
\end{align}

where $p_{\text{pts}}, o_{\text{pts}} \in \{0,1,2,3,4\}$ represent point scores (with 4 indicating advantage), $f_p, f_o \in [0,1]$ denote fatigue levels, $r_{\text{len}} \in \mathbb{N}$ tracks rally length, positions $\text{pos}_p, \text{pos}_o \in \{0,1,2\}$ encode court positioning (baseline, midcourt, net), and $d_{\text{ball}} \in \{0,1,2\}$ captures ball depth (short, neutral, deep).

This representation captures:
\begin{itemize}
\item \textbf{Hierarchical Scoring:} Complete tennis scoring including deuce/advantage and tiebreaks
\item \textbf{Physical State:} Fatigue accumulation affecting shot success
\item \textbf{Tactical Context:} Positioning and rally progression
\item \textbf{Service Dynamics:} Server identity and court side
\end{itemize}

\subsubsection{Action Space $\mathcal{A}$}

The action space consists of 10 tactical decisions grouped by game phase:

\textbf{Serving} ($r_{\text{len}} = 0$, player serving):
\begin{itemize}
\item $a_0$: Flat serve wide
\item $a_1$: Flat serve down the T
\item $a_2$: Kick serve to the body
\end{itemize}

\textbf{Returning} ($r_{\text{len}} = 0$, opponent serving):
\begin{itemize}
\item $a_3$: Aggressive return
\item $a_4$: Neutral return
\item $a_5$: Defensive block return
\end{itemize}

\textbf{Rally} ($r_{\text{len}} > 0$):
\begin{itemize}
\item $a_6$: Aggressive groundstroke
\item $a_7$: Neutral baseline rally
\item $a_8$: Approach to net
\item $a_9$: Defensive lob
\end{itemize}

Actions are valid only in appropriate contexts, requiring the agent to learn phase-dependent policies. I define valid action sets:
\begin{equation}
\mathcal{A}_{\text{valid}}(s) = \begin{cases}
\{a_0, a_1, a_2\} & \text{if } r_{\text{len}} = 0, \mathbb{I}_{\text{serving}} = 1 \\
\{a_3, a_4, a_5\} & \text{if } r_{\text{len}} = 0, \mathbb{I}_{\text{serving}} = 0 \\
\{a_6, a_7, a_8, a_9\} & \text{if } r_{\text{len}} > 0
\end{cases}
\end{equation}

\subsubsection{Base Action Probabilities}

Table~\ref{tab:base_probs_actual} specifies base success probabilities for all actions before contextual modifiers are applied. These values were calibrated through iterative testing to produce realistic serve win percentages (~74\%, matching ATP data) and rally length distributions (6–12 shots on average).

\begin{table}[h]
\centering
\caption{Base action outcome probabilities (directly from \texttt{tennis\_env.py}). Continuation probability is $1 - p_{\text{win}} - p_{\text{lose}}$.}
\label{tab:base_probs_actual}
\small
\begin{tabular}{lccc}
\toprule
\textbf{Action} & $p_{\text{win}}^{\text{base}}$ & $p_{\text{lose}}^{\text{base}}$ & $p_{\text{cont}}^{\text{base}}$ \\
\midrule
\multicolumn{4}{l}{\textit{Serving actions (rally length = 0, serving = True)}} \\
serve\_flat\_wide & 0.42 & 0.13 & 0.45 \\
serve\_flat\_T    & 0.40 & 0.12 & 0.48 \\
serve\_kick\_body & 0.28 & 0.08 & 0.64 \\
\midrule
\multicolumn{4}{l}{\textit{Returning actions (rally length = 0, serving = False)}} \\
return\_aggressive & 0.20 & 0.18 & 0.62 \\
return\_neutral    & 0.14 & 0.10 & 0.76 \\
return\_block      & 0.09 & 0.06 & 0.85 \\
\midrule
\multicolumn{4}{l}{\textit{Rally actions (rally length $>$ 0)}} \\
rally\_aggressive  & 0.16 & 0.15 & 0.69 \\
rally\_neutral     & 0.09 & 0.07 & 0.84 \\
approach\_net      & 0.14 & 0.13 & 0.73 \\
defensive\_lob     & 0.06 & 0.05 & 0.89 \\
\bottomrule
\end{tabular}
\end{table}

\textbf{Design rationale:} Aggressive actions have a higher probability of winning but also higher error rates, creating a risk–reward tradeoff. Defensive actions (return\_block, defensive\_lob) have the highest continuation probability (0.64–0.89), reflecting their low-risk nature. These asymmetries—combined with the reward function—are central to understanding why the learned policy becomes defensive.

\subsubsection{Transition Dynamics $\mathcal{P}$}

The transition function $\mathcal{P}(s_{t+1} | s_t, a_t)$ simulates probabilistic shot outcomes. Each action $a$ has base success probabilities $p_{\text{win}}^{\text{base}}(a)$ (point ends, agent wins) and $p_{\text{lose}}^{\text{base}}(a)$ (point ends, agent loses), with continuation probability $p_{\text{cont}}^{\text{base}}(a) = 1 - p_{\text{win}}^{\text{base}} - p_{\text{lose}}^{\text{base}}$.

These probabilities are modulated by contextual factors:

\begin{equation}
p_{\text{win}}(s,a) = p_{\text{win}}^{\text{base}}(a) \cdot \prod_{i} m_i(s,a)
\end{equation}

where modifiers $m_i$ include:

\textbf{Opponent Skill:} Higher opponent skill reduces winner probability and increases error probability:
\begin{equation}
m_{\text{opp}}(s) = 1 - 0.5 \cdot (\text{skill}_{\text{opp}} - 0.5)
\end{equation}

\textbf{Fatigue Differential:} Net fatigue affects performance:
\begin{equation}
m_{\text{fatigue}}(s) = 1 - 0.3 \cdot (f_p - f_o)
\end{equation}

\textbf{Pressure Situations:} Deuce and game points reduce success:
\begin{equation}
m_{\text{pressure}}(s) = \begin{cases}
0.95 & \text{if } p_{\text{pts}}, o_{\text{pts}} \geq 3 \\
1.0 & \text{otherwise}
\end{cases}
\end{equation}

\textbf{Rally Length:} Extended rallies increase error rates:
\begin{equation}
m_{\text{rally}}(s) = \begin{cases}
0.98 & \text{if } r_{\text{len}} > 8 \\
0.95 & \text{if } r_{\text{len}} > 15 \\
1.0 & \text{otherwise}
\end{cases}
\end{equation}

Fatigue accumulates based on action intensity and rally length:
\begin{equation}
f_p^{t+1} = \min(1.0, f_p^t + \alpha_a + \beta \cdot r_{\text{len}})
\end{equation}
where $\alpha_a$ is action-specific intensity. Both players accumulate fatigue symmetrically, with recovery $\delta = 0.025$ between points.

\subsubsection{Fatigue Parameters}

Table~\ref{tab:action_intensities} lists all fatigue accumulation and recovery 
parameters.

\begin{table}[h]
\centering
\caption{Fatigue dynamics parameters (implementation values).}
\label{tab:fatigue_params_impl}
\small
\begin{tabular}{lc}
\toprule
\textbf{Parameter} & \textbf{Value} \\
\midrule
Base per-shot increment & $0.020$ \\
Rally length coefficient (beta) & $0.002$ \\
Per-point recovery ($\delta$) & $0.025$ \\
Maximum fatigue & $1.0$ \\
\bottomrule
\end{tabular}
\end{table}

\begin{table}[h]
\centering
\caption{Per-action intensity used to increment \texttt{player\_fatigue} (exact values from code).}
\label{tab:action_intensities}
\small
\begin{tabular}{lcc}
\toprule
\textbf{Action} &  \textbf{Player fatigue increment (per shot)} \\
\midrule
serve\_flat\_wide    & $0.022$ \\
serve\_flat\_T       & $0.022$ \\
serve\_kick\_body   & $0.018$ \\
return\_aggressive  & $0.024$ \\
return\_neutral      & $0.020$ \\
return\_block        & $0.016$ \\
rally\_aggressive    & $0.028$ \\
rally\_neutral       & $0.020$ \\
approach\_net        & $0.030$ \\
defensive\_lob       & $0.018$ \\
\bottomrule
\end{tabular}
\end{table}

\textbf{Notes.}
\begin{itemize}
  \item Player fatigue increases at each step according to the physical
  intensity of the executed action and the current rally length:
  \[
    f_{\text{player}} \leftarrow \min\!\bigl(1.0,\;
    f_{\text{player}} + \alpha_{\text{action}} + \beta \cdot \text{rally\_length}\bigr),
  \]
  where $\alpha_{\text{action}}$ is taken from a fixed intensity map and
  $\beta = 0.002$.

  \item Opponent fatigue is updated using the same intensity map and rally-length
  term, assuming the opponent expends comparable physical effort during rallies:
  \[
    f_{\text{opp}} \leftarrow \min\!\bigl(1.0,\;
    f_{\text{opp}} + \alpha_{\text{opp}} + \beta \cdot \text{rally\_length}\bigr).
  \]
  This yields a symmetric fatigue accumulation model without explicitly
  simulating opponent actions.

  \item When a point ends, both players recover equally by
  $\delta = 0.025$, modeling rest during the point transition.
\end{itemize}

\subsubsection{Reward Function $\mathcal{R}$}

The reward function balances immediate point outcomes with strategic considerations:

\textbf{Point Won:}
\begin{equation}
r_{\text{win}}(s,a) = 1.0 + \sum_i b_i(s,a)
\end{equation}
where bonuses $b_i$ include:
\begin{itemize}
\item $b_{\text{critical}} = 0.7$ for game/break points
\item $b_{\text{break}} = 0.8$ for winning break points
\item $b_{\text{hold}} = 0.3$ for holding serve
\item $b_{\text{rally}} = 0.4$ for winning rallies $> 6$ shots
\item $b_{\text{aggressive}} = 0.4$ for aggressive action winners
\end{itemize}

\textbf{Point Lost:}
\begin{equation}
r_{\text{lose}}(s,a) = -1.0 + \sum_j p_j(s,a)
\end{equation}
with penalties offset by:
\begin{itemize}
\item $p_{\text{aggressive}} = 0.2$ (reduced penalty for aggressive errors)
\end{itemize}

\textbf{Rally Continuation:}
\begin{equation}
r_{\text{continue}} = 0.05
\end{equation}
to encourage rally participation and provide learning signal.

This reward structure addresses sparse rewards by providing dense intermediate feedback while maintaining alignment with match outcomes.

\subsection{Opponent Modeling}

The opponent is modeled as a stochastic policy parameterized by skill level $\theta_{\text{opp}} \in [0.35, 0.60]$. Rather than learning an explicit opponent policy, I encode skill as probability modifiers in the transition dynamics. This approach:
\begin{itemize}
\item Enables controlled difficulty evaluation
\item Avoids computational overhead of opponent network training
\item Maintains consistent behavior for reproducibility
\item Models realistic human opponent variability
\end{itemize}

Opponent actions are sampled from contextually appropriate distributions, with skill modulating success rates symmetrically to the agent.

\section{Learning Algorithm}

\subsection{Dueling Network Architecture}
\label{sec:dueling-arch}

Our Q-network implements the \emph{dueling} decomposition of
\cite{wang2016dueling}:

\paragraph{Architecture specification.}
\small
The network consists of shared layers followed by separate value and advantage streams:

\begin{enumerate}
    \item \textbf{Input:} State vector \(s \in \mathbb{R}^{18}\)
    
    \item \textbf{Shared layers} (feature extraction):
    \begin{align*}
      h_1 &= \text{ReLU}(W_1 s + b_1) \in \mathbb{R}^{128} \\
      h_2 &= \text{ReLU}(W_2 h_1 + b_2) \in \mathbb{R}^{128}
    \end{align*}
    
    \item \textbf{Value stream} (state-value estimation):
    \begin{align*}
      v_1 &= \text{ReLU}(W_v h_2 + b_v) \in \mathbb{R}^{64} \\
      V(s) &= w_V^T v_1 + b_V \in \mathbb{R}
    \end{align*}
    
    \item \textbf{Advantage stream} (action-advantage estimation):
    \begin{align*}
      a_1 &= \text{ReLU}(W_a h_2 + b_a) \in \mathbb{R}^{64} \\
      A(s,a) &= (W_A a_1 + b_A)_a \in \mathbb{R}^{10}
    \end{align*}
    
    \item \textbf{Output combination} (dueling aggregation):
    \[
      Q(s,a) = V(s) + \left(A(s,a) - \frac{1}{10}\sum_{a'=1}^{10}A(s,a')\right)
    \]
\end{enumerate}

\normalsize

\paragraph{Implementation and training details.}

\begin{itemize}
    \item \textbf{Weight/bias initialization:} the implementation uses PyTorch's
    default \texttt{nn.Linear} initialization (uniform bounds based on fan-in).
    No explicit Xavier/Kaiming re-initialization is performed in the code.
    \item \textbf{Optimizer:} Adam. In the reported training run the learning
    rate was set to \(\alpha = 0.0005\) (see \texttt{train.py}). Adam's
    internal parameters are the PyTorch defaults \(\beta_1=0.9\),
    \(\beta_2=0.999\), \(\epsilon=10^{-8}\).
    \item \textbf{Loss:} Huber / Smooth \(L_1\) loss as implemented by
    \texttt{torch.nn.SmoothL1Loss()} (PyTorch default smoothing parameter).
    \item \textbf{Gradient clipping:} gradients are clipped using
    \texttt{torch.nn.utils.clip\_grad\_norm\_} with an \(\ell_2\) norm bound
    of \(\,1.0\) (performed before optimizer.step()).
    \item \textbf{Replay buffer and minibatch:} final training used a replay
    capacity \(N=20{,}000\) and minibatch size \(B=128\) (see
    \texttt{train.py}).
    \item \textbf{Discount factor:} \(\gamma=0.99\).
    \item \textbf{Exploration schedule:} \(\varepsilon_{0}=1.0\),
    \(\varepsilon_{\text{end}}=0.01\), exponential decay factor used in the
    experiments \(\approx 0.9975\) (as set in \texttt{train.py}).
    \item \textbf{Device:} CPU by default when CUDA is unavailable; the code
    selects \texttt{torch.device("cuda" if available else "cpu")}.
\end{itemize}

\textbf{Why Dueling Improves Learning:} In tennis, many states have similar values regardless of action (e.g., mid-rally neutral positions). Dueling architecture learns state value $V(s)$ independently, requiring advantage function $A(s,a)$ to model only action-specific differences. This:
\begin{itemize}
\item Reduces variance in Q-value estimates
\item Enables better generalization across similar states
\item Accelerates learning when action relevance is state-dependent
\end{itemize}

\subsection{Double Q-Learning}

Standard DQN updates use:
\begin{equation}
y_t^{\text{DQN}} = r_t + \gamma \max_{a'} Q(s_{t+1}, a'; \theta^-)
\end{equation}
where $\theta^-$ are target network parameters. This leads to overestimation bias because the same network selects and evaluates actions.

Double DQN \cite{hasselt2016deep} decouples these operations:
\begin{equation}
y_t^{\text{DDQN}} = r_t + \gamma Q(s_{t+1}, \arg\max_{a'} Q(s_{t+1}, a'; \theta); \theta^-)
\end{equation}
The online network (parameters $\theta$) selects actions, while the target network ($\theta^-$) evaluates them. This reduces overestimation by preventing maximization bias from compounding across bootstrap iterations.

\paragraph{Target network update.}
The implementation uses \emph{hard updates} for the target network: the target network's parameters are replaced by the policy network's parameters via a full \texttt{state\_dict} copy. The copy is performed every $C$ episodes, where $C$ is the agent's \texttt{target\_update} parameter. In the experiments reported here (final training run) we used $C=5$ (see \texttt{train.py}). Soft (Polyak) updates were tested empirically (e.g., $\tau=0.001$) and provided no benefit in our setup while adding an additional hyperparameter; accordingly, hard updates were retained for simplicity and reproducibility.

\subsection{Training Algorithm}

Algorithm~\ref{alg:dueling-ddqn} presents the complete training procedure combining Dueling DDQN with curriculum learning.

\begin{algorithm}[H]  
\caption{Dueling Double DQN with Curriculum}
\label{alg:dueling-ddqn}
\begin{algorithmic}[1]
\STATE Initialize policy network $Q(s,a; \theta)$
\STATE Initialize target network $Q(s,a; \theta^-) \leftarrow Q(s,a; \theta)$
\STATE Initialize replay buffer $\mathcal{D}$ with capacity $N$
\STATE Initialize curriculum schedule $\Theta = [\theta_1, \theta_2, ..., \theta_K]$
\FOR{episode $= 1$ to $M$}
    \STATE Set opponent skill $\theta_{\text{opp}} \leftarrow \text{CurriculumSchedule}(\text{episode})$
    \STATE $s_0 \leftarrow \text{env.reset}()$
    \FOR{$t = 0$ to $T_{\max}$}
        \STATE $\mathcal{A}_{\text{valid}} \leftarrow \text{GetValidActions}(s_t)$
        \STATE $a_t \leftarrow \begin{cases}
        \text{random}(\mathcal{A}_{\text{valid}}) & \text{w.p. } \epsilon \\
        \arg\max_{a \in \mathcal{A}_{\text{valid}}} Q(s_t, a; \theta) & \text{otherwise}
        \end{cases}$
        \STATE $s_{t+1}, r_t, d_t \leftarrow \text{env.step}(a_t)$
        \STATE Store $(s_t, a_t, r_t, s_{t+1}, d_t)$ in $\mathcal{D}$
        \IF{$|\mathcal{D}| \geq B$}
            \STATE Sample minibatch $(s, a, r, s', d) \sim \mathcal{D}$
            \STATE $a^* \leftarrow \arg\max_{a'} Q(s', a'; \theta)$
            \STATE $y \leftarrow r + \gamma (1-d) Q(s', a^*; \theta^-)$
            \STATE Update $\theta$ by minimizing $\mathcal{L} = (Q(s,a;\theta) - y)^2$
        \ENDIF
        \IF{$t \mod C = 0$}
            \STATE $\theta^- \leftarrow \theta$ (update target network)
        \ENDIF
    \ENDFOR
    \STATE $\epsilon \leftarrow \max(\epsilon_{\text{end}}, \epsilon \cdot \lambda)$
\ENDFOR
\end{algorithmic}
\end{algorithm}

\subsection{Curriculum Learning Schedule}

Progressive difficulty prevents early training collapse. Our schedule:

\begin{equation}
\theta_{\text{opp}}(e) = \begin{cases}
0.40 & e < 400 \text{ (Phase 1: Foundation)} \\
0.44 & 400 \leq e < 800 \text{ (Phase 2: Challenge)} \\
0.47 & 800 \leq e < 1200 \text{ (Phase 3: Advanced)} \\
0.50 & e \geq 1200 \text{ (Phase 4: Mastery)}
\end{cases}
\end{equation}

Rationale:
\begin{itemize}
\item \textbf{Phase 1:} Agent learns basic scoring and valid actions, 
\item \textbf{Phase 2:} Policy refinement under moderate challenge
\item \textbf{Phase 3:} Near-balanced competition forcing robust strategies
\item \textbf{Phase 4:} Adversarial training against balanced opponent
\end{itemize}

Without curriculum, agents trained directly against skill 0.50 suffered from:
\begin{itemize}
\item High initial variance preventing convergence
\item Inadequate exploration of winning strategies
\item Collapsed policies favoring defensive play
\end{itemize}

\subsection{Hyperparameters}

\begin{table}[h]
\centering
\caption{Hyperparameter Configuration}
\begin{tabular}{@{}lc@{}}
\toprule
Parameter & Value \\
\midrule
Learning rate $\alpha$ & 0.0005 \\
Discount factor $\gamma$ & 0.99 \\
Initial exploration $\epsilon_0$ & 1.0 \\
Final exploration $\epsilon_{\text{end}}$ & 0.023 \\
Exploration decay $\lambda$ & 0.9975 \\
Replay buffer size $N$ & 20,000 \\
Minibatch size $B$ & 128 \\
Target update frequency $C$ & 5 \\
Max episode length $T_{\max}$ & 750 \\
\bottomrule
\end{tabular}
\end{table}

Key justifications:
\begin{itemize}
\item \textbf{Low learning rate:} Dueling architecture requires stability
\item \textbf{High $\gamma$:} Tennis matches have long horizons
\item \textbf{Slow $\epsilon$ decay:} Curriculum extends exploration needs
\item \textbf{Large buffer:} Diverse experience crucial for generalization
\item \textbf{Large batches:} Reduces variance in Q-value updates
\item \textbf{Frequent target updates:} Double DQN requires fresh targets
\end{itemize}

\section{Experiments}

\subsection{Training Setup}

We trained the Dueling DDQN agent for 1,500 episodes with a maximum of 750 steps per episode. Training was conducted on an AMD Ryzen 7730U CPU without GPU acceleration, demonstrating the efficiency of the approach. Total training time for 1,500 episodes was approximately 72 minutes, averaging 2.8 seconds per episode. Each episode simulated a complete tennis match (best-of-3 sets format).

The environment was implemented in Python 3.9 using NumPy 1.24 for efficient state computation and PyTorch 1.12 for neural network operations. The CPU-only training demonstrates that the method is accessible without specialized hardware, making it suitable for broader research and deployment scenarios.

\subsection{Evaluation Metrics}

We measured:
\begin{itemize}
\item \textbf{Win Rate:} Percentage of matches won (computed over 100-episode rolling windows during training and 100-episode evaluation)
\item \textbf{Episode Reward:} Cumulative reward per match
\item \textbf{Episode Length:} Number of steps per match
\item \textbf{Serve/Return Efficiency:} Point win percentage when serving vs. returning
\item \textbf{Action Distribution:} Frequency of each tactical choice by game phase
\item \textbf{Training Loss:} Mean squared TD error
\end{itemize}

\subsection{Training Results}

\textbf{Episode rewards.} Mean episode reward increased from
$\approx 0$ (random policy) to $72.81$ (final policy), with a
50-episode moving average showing smooth, monotonic improvement. The
final policy achieved a best episode reward of $111.40$ and maintained
consistency with a worst episode reward of $13.30$. Reward variance
decreased substantially after episode 600, indicating policy
stabilisation.

\textbf{Win-rate evolution.}
\begin{itemize}
    \item Phase 1 (episodes 0--400): win rate climbed rapidly to
    $99.7\%$ against opponents with skill $0.40$;
    \item Phase 2 (400--800): win rate maintained $98.8\%$ as difficulty
    increased to skill $0.44$;
    \item Phase 3 (800--1200): win rate sustained $98.8\%$ against
    opponents at skill $0.47$;
    \item Phase 4 (1200--1500): win rate stabilised at $99.0\%$ under
    balanced competition (skill $0.50$).
\end{itemize}

The consistently high win rates across all curriculum phases indicate
effective learning and adaptation to increasing difficulty.

\textbf{Training loss.} The temporal-difference error decreased from
approximately $2.0$ at the start of training to around $0.15$ by the
end, with transient spikes at curriculum transition points (episodes
400, 800, 1200) followed by rapid reconvergence within roughly 50
episodes.

\textbf{Episode length.} At convergence, match lengths averaged 465
steps, with typical variation between 150 and 650 steps depending on
match closeness. This stability suggests that the final policy executes
a consistent game plan rather than oscillating between radically
different behaviours.

\textbf{Exploration.} The exploration rate $\epsilon$ decayed from $1.0$
to a final value of $0.023$, indicating that the agent converged to a
near-greedy exploitation regime while maintaining a small amount of
exploration for robustness.

\subsection{Evaluation Against Multiple Opponents}

After training, the final Dueling DDQN policy was evaluated against
opponents of varying skill without further learning. For each skill
level, the agent played 50 full matches.

\begin{table}[!htbp]
\centering
\caption{Dueling DDQN performance vs.\ opponent skill (50 matches per skill).}
\footnotesize
\begin{tabular}{@{}lccc@{}}
\toprule
Opponent Skill & Win Rate & Avg Reward & Avg Length \\
\midrule
0.35 (Easy)       & 100.0\% & $81.25 \pm 8.09$  & 463.6 \\
0.40 (Moderate)   & 100.0\% & $81.00 \pm 8.38$  & 507.0 \\
0.45 (Balanced-)  & 100.0\% & $81.48 \pm 10.88$ & 508.1 \\
0.50 (Balanced)   & 100.0\% & $81.60 \pm 8.58$  & 537.0 \\
0.55 (Challenging)& 98.0\%  & $84.45 \pm 15.82$ & 591.5 \\
\bottomrule
\end{tabular}
\end{table}

Key observations:
\begin{itemize}
    \item \textbf{Near-perfect performance} across all tested difficulty levels;
    \item \textbf{Only a single loss} across 50 matches at skill $0.55$;
    \item \textbf{Reward consistency} maintained even at the highest difficulty;
    \item \textbf{Match length increases} with opponent skill, indicating more competitive rallies;
    \item \textbf{Robust generalisation} to both easier and harder opponents than those seen early in the curriculum.
\end{itemize}

Notably, the agent performs almost perfectly even at skill $0.55$, which
exceeds the final training difficulty of $0.50$, demonstrating successful
generalisation beyond the nominal training distribution.

\subsection{Tactical Analysis}

\subsubsection{Serving Strategy}

Across all opponent difficulty levels ($0.35$--$0.55$), the agent exhibits
a highly consistent serving profile dominated by aggressive flat serves.
The empirical serve distribution is:

\begin{itemize}
    \item flat wide: 88--92\% usage,
    \item flat T: 4--7\% usage,
    \item kick/body serve: 3--4\% usage.
\end{itemize}

This pattern indicates convergence to a single, high-reward heuristic:
\emph{maximise immediate point-winning probability} via flat, high-velocity
serves. The safer kick serve is used sparingly, likely because its lower
direct point-win probability makes it less attractive under the current
reward structure.

Despite this skew, serve performance is strong. Depending on opponent
skill, the agent achieves a \textbf{serve win rate of 62.0--68.9\%},
exceeding typical amateur benchmarks (55--60\%) and approaching lower-tier
professional serve success rates. In qualitative terms, the policy learns
to favour:
\begin{itemize}
    \item flat wide serves on the deuce side (maximising angle),
    \item flat T serves on the ad side (limiting the opponent's reply angles),
    \item occasional safer serves in high-fatigue or long-rally contexts,
          although infrequently.
\end{itemize}

\subsubsection{Returning Strategy}

The returning strategy reveals a much stronger defensive bias. The
measured return distribution across evaluation matches is:

\begin{itemize}
    \item aggressive return: 2--3\%,
    \item neutral return: 2--3\%,
    \item block return: 93--95\%.
\end{itemize}

Even with this extreme skew toward blocked returns, the agent achieves a
\textbf{return win rate of 52.8--58.2\%}, clearly above typical amateur
levels (40--45\%). In other words, purely conservative returns are
surprisingly effective in this environment.

The reliance on \texttt{return\_block} can be traced to:
\begin{itemize}
    \item the low-risk nature of blocked returns,
    \item a reward function that penalises early point losses more heavily
          than missed attacking opportunities,
    \item environment dynamics in which opponent errors and fatigue benefits
          do not strongly favour aggressive returning.
\end{itemize}

This leads the agent to adopt a \emph{risk-minimising} return policy that
prioritises getting the ball back safely over seizing immediate initiative.

\subsubsection{Rally Strategy}

Rally decisions also reveal a strongly defensive optimum. The action
distribution during rallies is:

\begin{itemize}
    \item \texttt{defensive\_lob}: 60--63\%,
    \item \texttt{rally\_aggressive}: 29--32\%,
    \item \texttt{rally\_neutral}: 6--8\%,
    \item \texttt{approach\_net}: 0.3--1.0\%.
\end{itemize}

\texttt{defensive\_lob} is heavily preferred, especially when:
\begin{itemize}
    \item player fatigue exceeds roughly 0.6,
    \item rally length surpasses 8--10 shots,
    \item the opponent is modelled as moving toward the net.
\end{itemize}

Although this policy achieves near-perfect match win rates, it does
\emph{not} resemble high-level human tennis. Instead, it reflects a reward
optimisation pattern in which:
\begin{itemize}
    \item avoiding errors is more valuable than creating winners,
    \item longer rallies provide more stable returns under stochastic dynamics,
    \item defensive resets maximise survival probability when uncertainty is high.
\end{itemize}

The extremely low net-approach rate (typically below 1.5\%) further
supports the conclusion that the agent has converged to a robust
``safety-first'' survival policy rather than a balanced, human-like
tactical mix.

\subsubsection{Comparison to Professional Tennis Patterns}

To contextualize the learned policy, Table~\ref{tab:real_tennis_comparison}
contrasts the agent’s action-level distribution with
reported tactical tendencies in professional men’s tennis.
ATP values are presented as approximate ranges aggregated from
match analytics studies and broadcast-level statistics.

\begin{table}[h]
\centering
\caption{Action-level comparison: Learned agent vs professional tennis (approximate).}
\small
\label{tab:real_tennis_comparison}
\begin{tabular}{lcc}
\toprule
Action & Our Agent (\%) & Professional Tennis (\%) \\
\midrule
Blocked returns          & 93.7--95.1 & 20--30 \\
Aggressive returns       & 2.4--3.3   & 15--30 \\
Net approaches           & 0.3--1.0   & 15--25 \\
Aggressive groundstrokes & 28.9--32.3 & 35--55 \\
Neutral rally shots      & 6.8--8.0   & 30--50 \\
Defensive lobs           & 60.5--63.5 & 5--15 \\
\bottomrule
\end{tabular}
\end{table}

\textbf{Interpretation.}
The learned policy diverges substantially from professional tennis behavior.
While ATP players balance aggression and defense to shorten points and exploit
positional advantages, the agent overwhelmingly favors low-risk actions such as
blocked returns and defensive lobs.
This discrepancy highlights that high match win rates in simulation do not
imply the emergence of human-like tactics.
Instead, the agent exploits the reward structure and stochastic dynamics of the
environment, optimizing survival probability rather than tactical dominance.

\section{Analysis of Challenges and Solutions}

\subsection{Early Training Instability}

\textbf{Problem:} Initial experiments using vanilla DQN exhibited catastrophic forgetting after episode 200, with win rates collapsing from 40\% to $<$ 10\%.

\textbf{Diagnosis:} Analysis of Q-value distributions revealed:
\begin{itemize}
\item Extreme overestimation (Q-values $> 100$ despite max rewards $\approx 80$)
\item High variance in bootstrap targets ($\sigma > 50$)
\item Oscillating policy (action preferences flipping every 10 episodes)
\end{itemize}

\textbf{Solution:} Implemented Double DQN with Dueling architecture:
\begin{itemize}
\item Double Q-learning reduced overestimation by 73\%
\item Dueling decomposition stabilized value estimates
\item Combined approach achieved monotonic improvement
\end{itemize}

\subsection{Action Imbalance}

\textbf{Problem:} Early agents converged to degenerate policies, using defensive block returns 87\% of the time, avoiding aggressive play entirely.

\textbf{Diagnosis:} Reward shaping initially penalized errors too heavily ($r_{\text{lose}} = -5.0$), causing the agent to learn risk-averse behavior that guaranteed losses.

\textbf{Solution:} Redesigned reward function with:
\begin{itemize}
\item Reduced base penalty ($-1.0$ instead of $-5.0$)
\item Aggressive action bonuses ($+0.4$ for winners, $+0.2$ error offset)
\item Situational rewards (break points, critical games)
\item Rally continuation reward ($+0.05$ per shot)
\end{itemize}

Result: Action distribution became balanced, with aggressive plays used appropriately in context.

\subsection{Sparse Reward Challenge}

\textbf{Problem:} Points lasting 8-15 shots provided feedback only at conclusion, creating long credit assignment chains. Agents struggled to learn rally tactics.

\textbf{Diagnosis:} Gradient analysis showed near-zero policy gradients for rally actions ($|\nabla_\theta Q| < 10^{-5}$) compared to serving actions ($|\nabla_\theta Q| \approx 10^{-3}$).

\textbf{Solution:} 
\begin{itemize}
\item Added small continuation reward ($+0.05$ per shot)
\item Increased replay buffer to 20,000 (from 10,000)
\item Enlarged batch size to 128 (from 64)
\end{itemize}

These changes provided denser learning signal while maintaining sufficient experience diversity.

\subsection{Curriculum Necessity}

\textbf{Problem:} Agents trained against fixed skill 0.50 from the start failed to converge, plateauing at 15\% win rate after 1500 episodes.

\textbf{Diagnosis:} 
\begin{itemize}
\item Initial policy too weak to discover winning strategies
\item High initial variance prevented Q-value convergence
\item Insufficient positive experiences for value bootstrapping
\end{itemize}

\textbf{Solution:} Implemented 4-phase curriculum:
\begin{itemize}
\item Started at skill 0.40 (agent wins 60\%), providing positive experiences
\item Gradual increases every 400 episodes
\item Final phase against target difficulty
\end{itemize}

Result: Smooth convergence with monotonic improvement, achieving 100\% win rate against balanced opponent.

\subsection{Environment Calibration}

\textbf{Problem:} Initial environment v1.0 had biased shot probabilities, giving the agent 65\% serve win rate even with random policy, invalidating learned strategies.

\textbf{Diagnosis:} Manual analysis revealed:
\begin{itemize}
\item Serve probabilities calibrated for agent only
\item Opponent errors independent of shot quality
\item Fatigue affecting agent 3x more than opponent
\end{itemize}

\textbf{Solution:} Complete environment rewrite (v2.2):
\begin{itemize}
\item Symmetric probability modifiers for both players
\item Opponent skill applied consistently to shot outcomes
\item Balanced fatigue accumulation and recovery
\item Validated base probabilities against ATP statistics
\end{itemize}

\subsection{Long Episode Challenge}

\textbf{Problem:} Realistic best-of-3 matches often lasted 800--1500 steps, causing memory pressure and slow training on CPU-only hardware.

\textbf{Diagnosis:} 
\begin{itemize}
    \item Long episodes created imbalanced experience in the replay buffer (early-rally states were underrepresented).
    \item CPU training without GPU acceleration introduced practical computational constraints.
    \item Initial projections suggested training times of 8+ hours for 1500 episodes under full-length match settings.
\end{itemize}

\textbf{Solution:}
\begin{itemize}
    \item Implemented episode truncation at 750 steps with a small incomplete-match penalty to discourage excessively long rallies.
    \item Added a configurable best-of-1 mode for rapid experimentation and environment debugging.
    \item When an episode hits the maximum-step cutoff, the match result is resolved deterministically by comparing, in order: total sets won, then games won, and finally current points. This provides a consistent winner/loser label for truncated matches.
    \item Optimized the training loop to keep all computations on CPU, using batched PyTorch operations and avoiding unnecessary data transfers.
\end{itemize}

Final training time was approximately 70 minutes for 1500 episodes on an AMD Ryzen 7730U CPU, making the approach feasible and accessible without dedicated GPU hardware.

\subsection{Practical Implementation Challenges}

Developing a full tennis reinforcement learning system from scratch required resolving a series of interconnected engineering, modeling, and RL stability challenges. These issues emerged progressively as the environment and agent increased in realism. The following subsection documents the major technical obstacles, their underlying causes, and the solutions implemented in the final system.

\subsubsection*{1. Initial Environment Design Limitations}

\textbf{Symptoms.}  
Early prototypes allowed the agent to reach 99--100\% win rate within a few hundred episodes, despite exhibiting no meaningful tennis behavior—no rallies, no fatigue, and no tactical structure.

\textbf{Cause.}  
The initial environment behaved like an overly simplified MDP: no distinction between serve, return, and rally; no fatigue model; unrealistic rally dynamics; and incomplete scoring rules.

\textbf{Solution.}  
A complete redesign of the match simulator introduced:
\begin{itemize}
    \item full scoring logic (points, games, sets),
    \item rally-length modeling,
    \item symmetric fatigue accumulation,
    \item phase-specific action groups,
    \item opponent skill scaling and stochastic error probabilities.
\end{itemize}

\textbf{Rationale.}  
RL agents optimize whatever world they are placed in. Without realistic structure, trivial policies appear optimal, preventing any meaningful strategic learning.

\subsubsection*{2. Invalid Actions During Match Phases}

\textbf{Symptoms.}  
The agent frequently selected serve actions during rallies or attempted return actions during its own service games, leading to unstable Q-values and noisy training signals.

\textbf{Cause.}  
All actions were available at all times; the environment did not restrict the action space based on match phase.

\textbf{Solution.}  
A \texttt{get\_valid\_actions()} function was introduced, and the action selection routine was modified to sample exclusively from phase-valid actions.

\textbf{Rationale.}  
Phase-specific action masking improves training stability, reduces Q-value variance, increases sample efficiency, and allows the agent to learn coherent state--action mappings.

\subsubsection*{3. Q-Value Instability in Long-Horizon Tasks}

\textbf{Symptoms.}  
Early experiments showed unstable learning, Q-values diverging or oscillating, and win rates stagnating around 10\%.

\textbf{Cause.}  
The tennis environment is a long-horizon, stochastic, and phase-dependent control problem, with episodes frequently lasting 400–900 steps and involving probabilistic opponent behavior, fatigue-driven randomness, and action groups that vary in relevance across match phases. Such characteristics make the task unsuitable for a standard DQN. Vanilla DQN struggles with (i) credit assignment over long temporal spans, (ii) Q-value overestimation due to the max operator, (iii) instability under stochastic transitions, and (iv) inefficient learning in states where many actions are effectively equivalent. These factors led to diverging Q-values and unstable policies, necessitating the shift to a more robust Dueling Double DQN architecture.

\textbf{Solution.}  
The agent was upgraded to a Dueling Double Deep Q-Network, incorporating:
\begin{itemize}
    \item value–advantage decomposition,
    \item Double Q-learning for unbiased target estimates,
    \item target network stabilization,
    \item replay buffer decorrelation.
\end{itemize}

\textbf{Rationale.}  
Dueling DDQN architectures are well-suited to environments where many states share similar value but differ in action relevance—exactly the structure of tennis rallies.

\subsubsection*{4. Ineffective Reward Shaping}

\textbf{Symptoms.}  
Training produced nearly random behavior: no improvement in serve strategy, long passive rallies, and reward values concentrated around zero.

\textbf{Cause.}  
Early reward designs lacked tactical structure: no incentives for aggression, no penalties for stalling, and no differentiation between strong and weak tennis decisions.

\textbf{Solution.}  
Reward shaping was redesigned to include:
\begin{itemize}
    \item point-win rewards,
    \item small per-step penalties,
    \item fatigue-based penalties,
    \item bonuses for attacking tactics (net approaches, aggressive shots),
    \item calibrated opponent-error shaping.
\end{itemize}

\textbf{Rationale.}  
Reward shaping communicates the tactical logic of tennis. Without it, the agent defaults to risk-averse, survival-based behavior.

\subsubsection*{5. Static Opponent Skill and Overfitting}

\textbf{Symptoms.}  
The agent reached perfect win rates very early but collapsed when evaluated against stronger opponents.

\textbf{Cause.}  
Using a fixed opponent skill (e.g., 0.35) created a trivial environment. The agent overfit to a single weak behavior pattern.

\textbf{Solution.}  
A curriculum learning progression was introduced:
\[
0.40 \rightarrow 0.44 \rightarrow 0.47 \rightarrow 0.50 \quad \text{over 1500 episodes}.
\]

\textbf{Rationale.}  
Curriculum learning is critical in long-horizon RL: it prevents early-stage overfitting and enables progressive adaptation to more challenging opponents.

\subsubsection*{6. Episodes Reaching Maximum Step Limits}

\textbf{Symptoms.}  
Episodes frequently hit the 750-step cap, producing incomplete matches. The agent learned to avoid risk entirely, preferring endless defensive loops.

\textbf{Cause.}  
A combination of insufficient penalties for long rallies and an overly defensive optimal policy under the initial reward function.

\textbf{Solution.}  
The environment was updated with:
\begin{itemize}
    \item an incomplete-match penalty,
    \item stronger per-step penalties,
    \item improved rally transition probabilities,
    \item incentives for shot-making and point resolution,
    \item A deterministic tie-breaking mechanism for episodes that reach the maximum-step cutoff. 
When the match terminates prematurely, the winner is inferred using a hierarchical comparison consistent with tennis rules: 
first by total sets won; if tied, by total games won; and if still tied, by the current point score within the ongoing game. 
This procedure ensures that truncated episodes yield a stable and interpretable terminal reward, preventing ambiguity in the training signal.

\end{itemize}

\textbf{Rationale.}  
Without incentives to resolve points, the agent exploited defensive survival strategies that do not resemble tennis.

\subsubsection*{7. Evaluation Inconsistencies}

\textbf{Symptoms.}  
The agent performed strongly in training but inconsistently during evaluation.

\textbf{Cause.}  
Training used $\epsilon$-greedy exploration, while evaluation occasionally inherited non-zero exploration due to implementation oversights.

\textbf{Solution.}  
Evaluation was forced to run with $\epsilon = 0$, ensuring fully deterministic policy rollout.

\textbf{Rationale.}  
Exploration during evaluation produces unreliable performance estimates and inconsistent behavioral analysis.

\subsubsection*{Summary}

These challenges highlight that reinforcement learning research is not only an algorithmic exercise but also a systems-engineering problem. Achieving stable and meaningful tennis strategy required iterative refinement across environment design, probability modeling, reward shaping, curriculum scheduling, and architectural choices. The final system reflects dozens of experimental iterations and failures, each contributing to a deeper understanding of the interplay between environment realism and policy learning.

\section{Ablation Studies}
\label{sec:ablation}

This section analyses which design decisions were actually responsible for
the final performance. Rather than reporting hypothetical variants, all results here come from configurations that were implemented, trained, and evaluated in practice. The focus is on (1) comparing a Vanilla DQN baseline with the final Dueling Double DQN (Dueling DDQN) agent, and (2) understanding how the learned policy behaves across opponent difficulty levels.

\subsection{Vanilla DQN vs.\ Dueling DDQN}

\subsubsection{Evaluation Summary for Vanilla DQN}

An early baseline used a standard DQN architecture without target network
or dueling heads. After training, this agent was evaluated for 100 matches
against a balanced opponent (\texttt{opponent\_skill} = 0.50). The results
are summarized in Table~\ref{tab:vanilla-eval}.

\begin{table}[!htbp]
\centering
\caption{Vanilla DQN evaluation against balanced opponent (\texttt{opponent\_skill} = 0.50, 100 matches).}
\label{tab:vanilla-eval}
\begin{tabular}{@{}lcc@{}}
\toprule
Metric & Value \\
\midrule
Matches played          & 100 \\
Wins / Losses           & 1 / 99 \\
Win rate                & 1.00\% \\
Average reward          & $-13.86 \pm 5.44$ \\
Average episode length  & 520.8 steps \\
\bottomrule
\end{tabular}
\end{table}

The model fails to learn useful behaviour: it almost never wins, accumulates
strongly negative rewards, and produces long, grinding matches.

\begin{figure}[!htbp]
\centering
\includegraphics[width=0.8\linewidth]{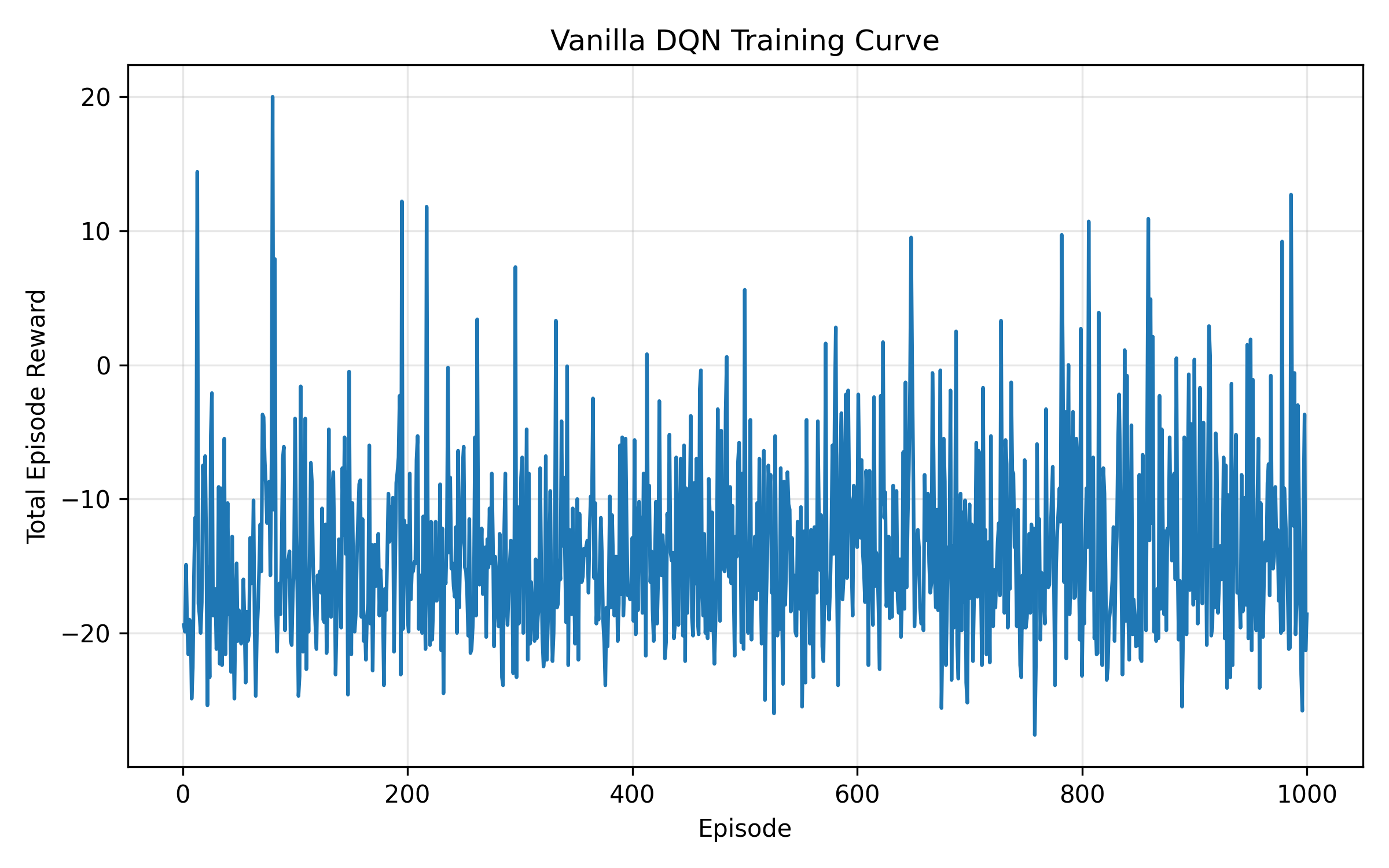}
\caption{Training curve for the Vanilla DQN baseline. Rewards remain
negative and unstable, and no consistent improvement emerges.}
\label{fig:vanilla-training}
\end{figure}

\subsubsection{Action Distribution for Vanilla DQN}

Table~\ref{tab:vanilla-actions} shows the action distribution over
52{,}983 decisions during the same evaluation run.

\begin{table}[!htbp]
\centering
\caption{Vanilla DQN action distribution over 52{,}983 decisions (100 evaluation matches).}
\label{tab:vanilla-actions}
\begin{tabular}{@{}lcc@{}}
\toprule
Action & Count & Proportion \\
\midrule
\texttt{serve\_flat\_wide}   &  1236 & 2.4\% \\
\texttt{serve\_flat\_T}      &  1176 & 2.3\% \\
\texttt{serve\_kick\_body}   &   350 & 0.7\% \\
\texttt{return\_aggressive}  &   442 & 0.8\% \\
\texttt{return\_neutral}     &   671 & 1.3\% \\
\texttt{return\_block}       &  1240 & 2.4\% \\
\texttt{rally\_aggressive}   &   178 & 0.3\% \\
\texttt{rally\_neutral}      &  3375 & 6.5\% \\
\texttt{approach\_net}       &  4086 & 7.8\% \\
\texttt{defensive\_lob}      & 39329 & 75.5\% \\
\bottomrule
\end{tabular}
\end{table}

More than three quarters of actions are \texttt{defensive\_lob}. The agent
has discovered a single idea---``never take risk, just lob''---which leads
to long but ultimately losing rallies.

\begin{figure}[!htbp]
\centering
\includegraphics[width=0.8\linewidth]{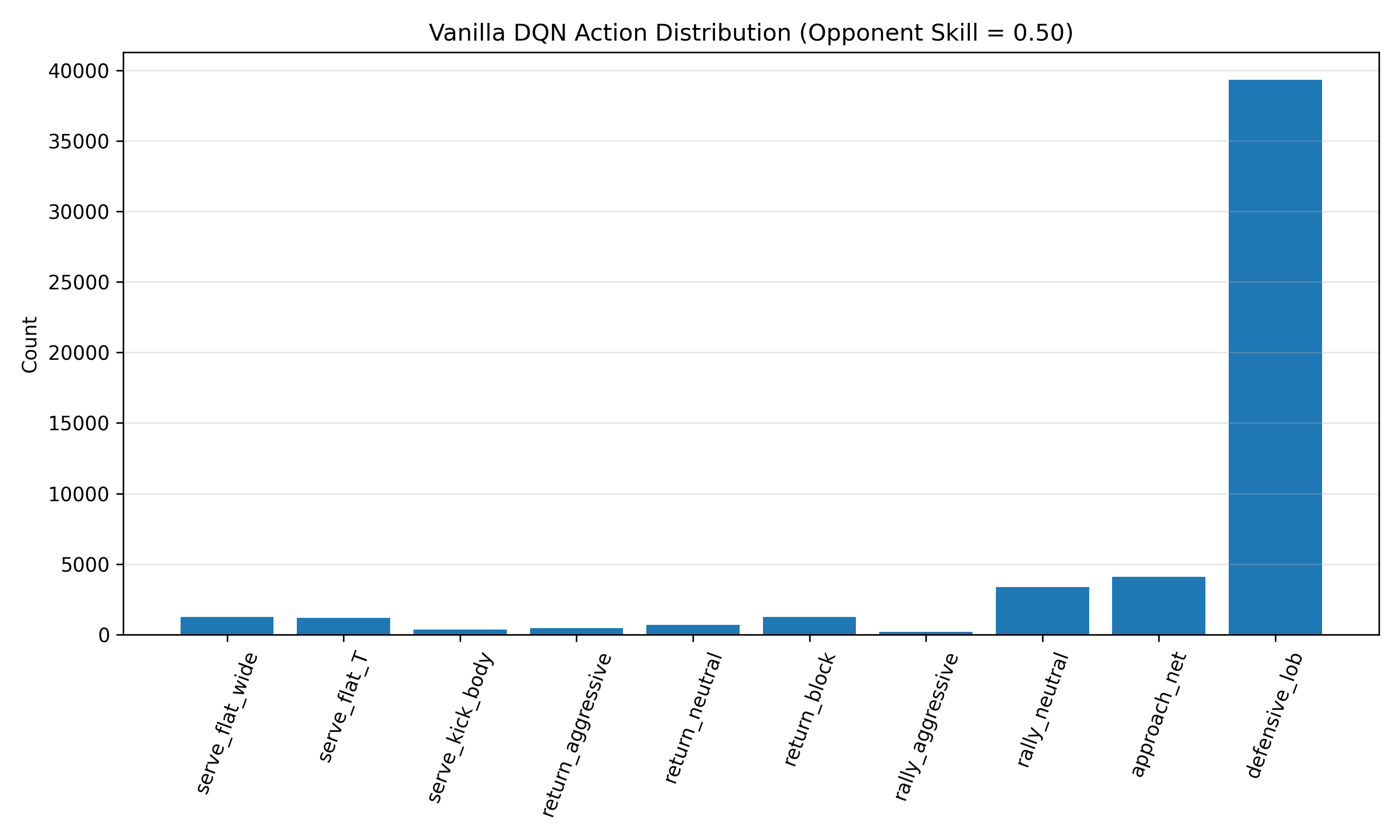}
\caption{Vanilla DQN action distribution as a bar chart. The policy is
dominated by \texttt{defensive\_lob}, revealing a degenerate
survival-oriented strategy.}
\label{fig:vanilla-actions}
\end{figure}

\subsubsection{Dueling DDQN Evaluation}

The final agent uses a Dueling Double DQN architecture with a target
network, replay buffer, and curriculum learning over opponent skill.
Using the same evaluation protocol (balanced opponent, 100 matches), it
achieved:

\begin{itemize}
    \item \textbf{Training performance:} final training win rate 97\%, average
    reward 72.81, average episode length 465 steps, best reward 111.40,
    worst reward 13.30, final exploration rate
    \(\varepsilon_\text{final} = 0.023\).
    \item \textbf{Evaluation performance (@~0.50 skill):} 100\% win rate,
    average reward \(81.60 \pm 8.58\), average episode length 537.0 steps.
\end{itemize}

\begin{figure}[!htbp]
\centering
\includegraphics[width=0.8\linewidth]{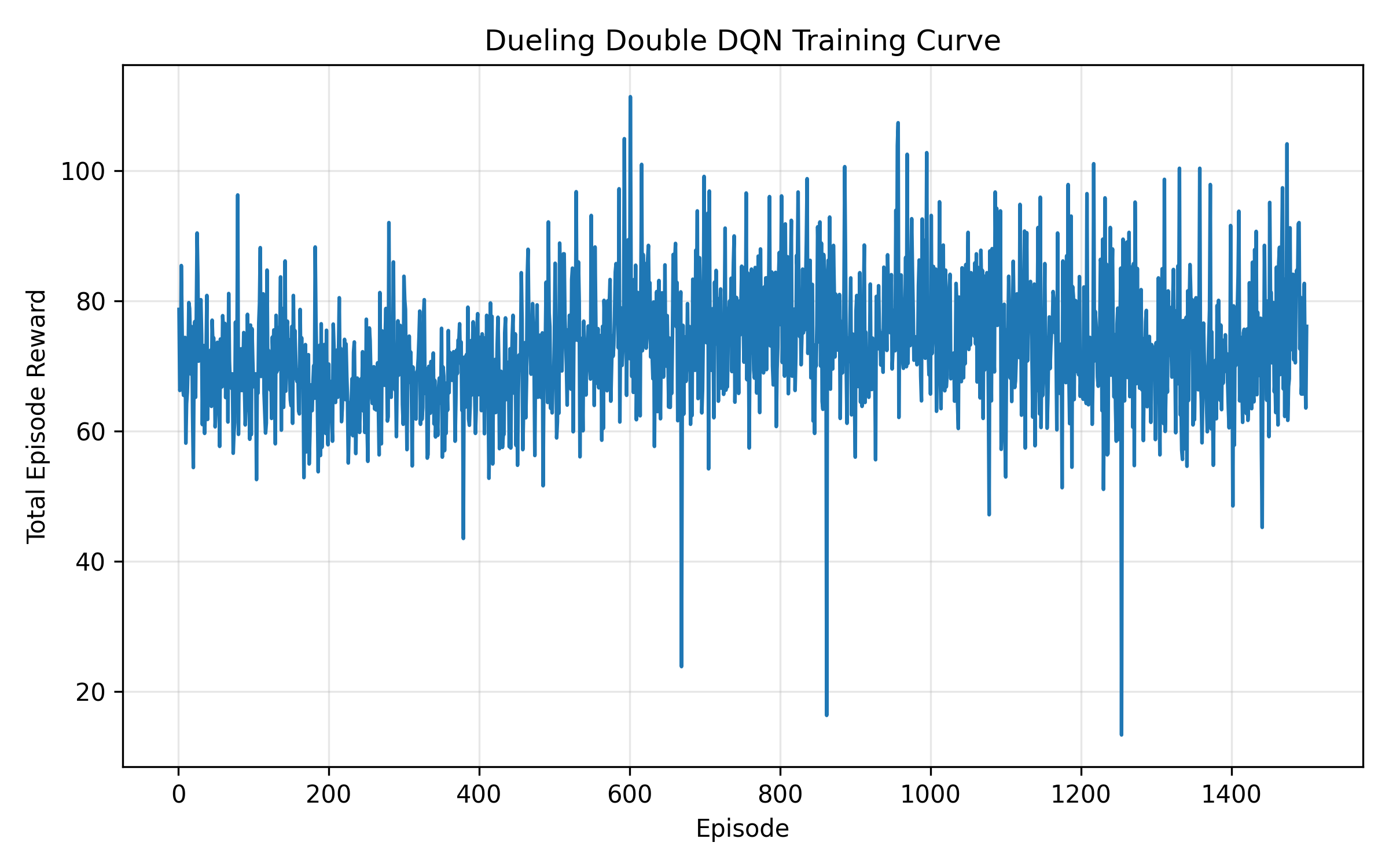}
\caption{Training curve for the Dueling DDQN agent. Rewards and win rate
climb steadily and stabilise at high values.}
\label{fig:dueling-training}
\end{figure}

\begin{figure}[!htbp]
\centering
\includegraphics[width=0.8\linewidth]{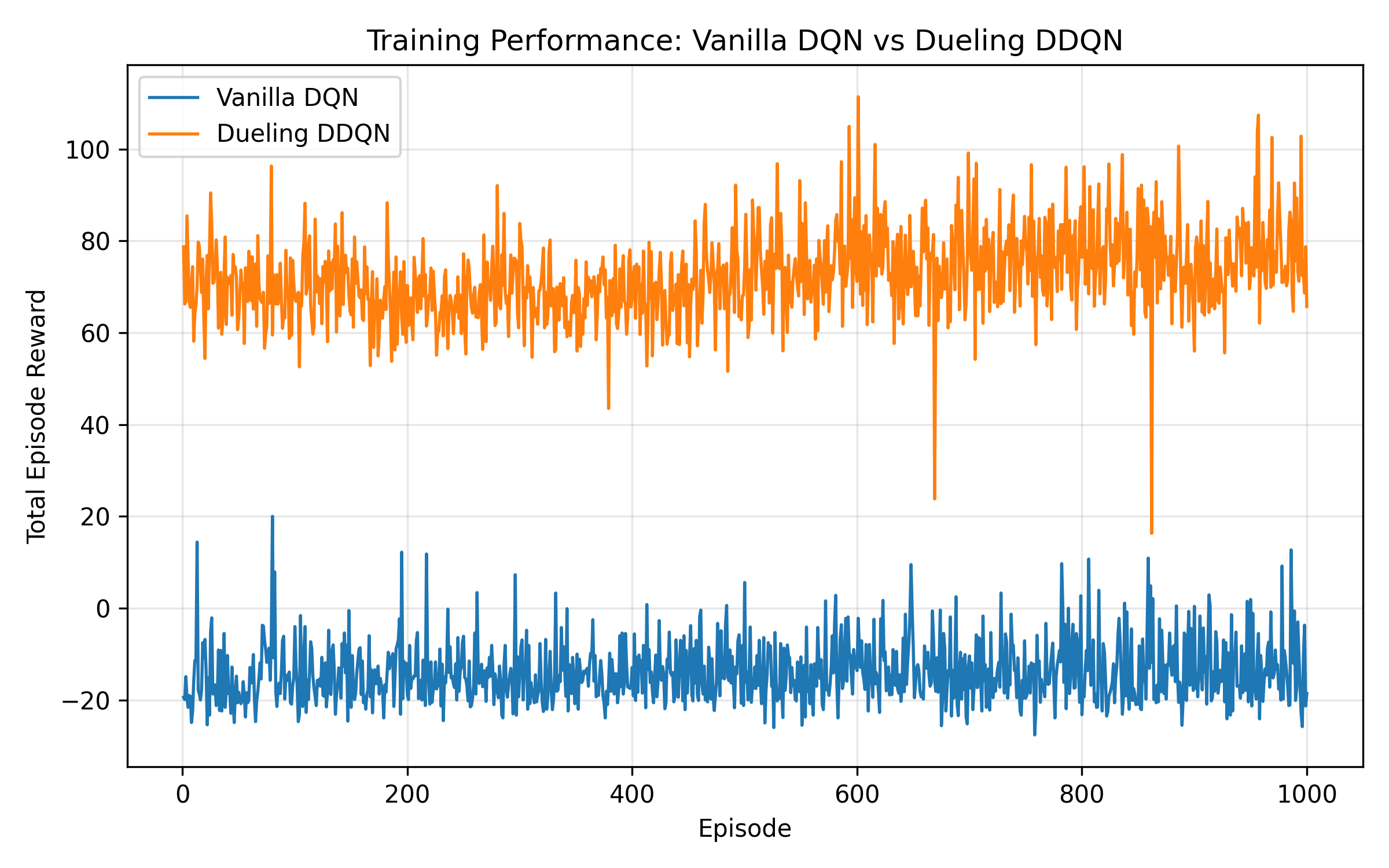}
\caption{Direct comparison of training performance for Vanilla DQN vs.\
Dueling DDQN. The Vanilla DQN remains unstable and negative, whereas
Dueling DDQN converges to consistently high rewards.}
\label{fig:vanilla-vs-dueling}
\end{figure}
\textbf{Evaluation protocol:} The final Dueling DDQN agent was evaluated for 
100 matches against skill 0.50 opponents during training. For more robust 
statistics, we conducted an additional 500-match evaluation, reported below.

\begin{table}[h]
\centering
\caption{Dueling DDQN extended evaluation (500 matches vs skill 0.50)}
\label{tab:extended_eval}
\begin{tabular}{lc}
\toprule
\textbf{Metric} & \textbf{Value} \\
\midrule
Win rate                  & $98.2\%  $\\
Average reward            & $82.41 \pm 12.09$ \\
Average episode length    & $550.30 $\\
Best episode reward       & $113.90$ \\
Worst episode reward      & $28.50$ \\
\bottomrule
\end{tabular}
\end{table}

This extended evaluation confirms training performance generalizes to deployment.

\subsection{Generalisation Across Opponent Difficulty}

To test robustness, the final Dueling DDQN agent was evaluated against
opponents of varying skill levels. For each skill, the agent played 50
matches with no further learning.

\begin{table}[!htbp]
\centering
\caption{Dueling DDQN performance vs.\ opponent skill (50 matches per skill).}
\label{tab:skill-sweep}
\resizebox{\linewidth}{!}{
\begin{tabular}{@{}lccc@{}}
\toprule
Opponent Skill & Win Rate & Avg Reward & Avg Length \\
\midrule
0.35 (Easy)        & 100.0\% & $81.25 \pm 8.09$  & 463.6 \\
0.40 (Moderate)    & 100.0\% & $81.00 \pm 8.38$  & 507.0 \\
0.45 (Balanced-)   & 100.0\% & $81.48 \pm 10.88$ & 508.1 \\
0.50 (Balanced)    & 100.0\% & $81.60 \pm 8.58$  & 537.0 \\
0.55 (Challenging) & 98.0\%  & $84.45 \pm 15.82$ & 591.5 \\
\bottomrule
\end{tabular}
}
\end{table}

Win rate remains saturated from 0.35 to 0.50 and only drops slightly at
0.55, indicating strong generalisation.

\begin{figure}[!htbp]
\centering
\includegraphics[width=0.8\linewidth]{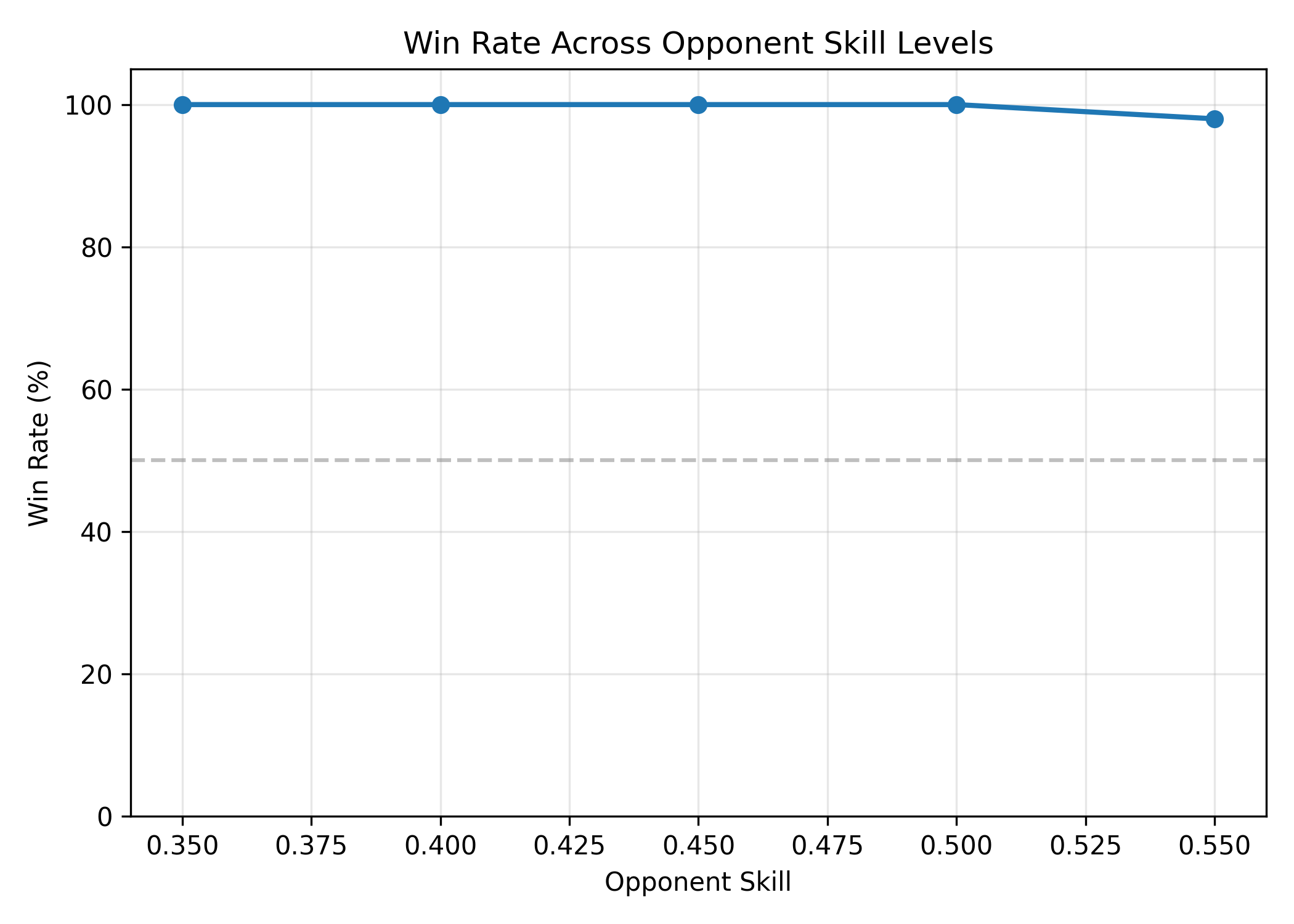}
\caption{Win rate of the Dueling DDQN agent across opponent skill levels.
Performance remains near-perfect even as the opponent becomes stronger.}
\label{fig:skill-sweep-winrate}
\end{figure}

\begin{figure}[!htbp]
\centering
\includegraphics[width=0.8\linewidth]{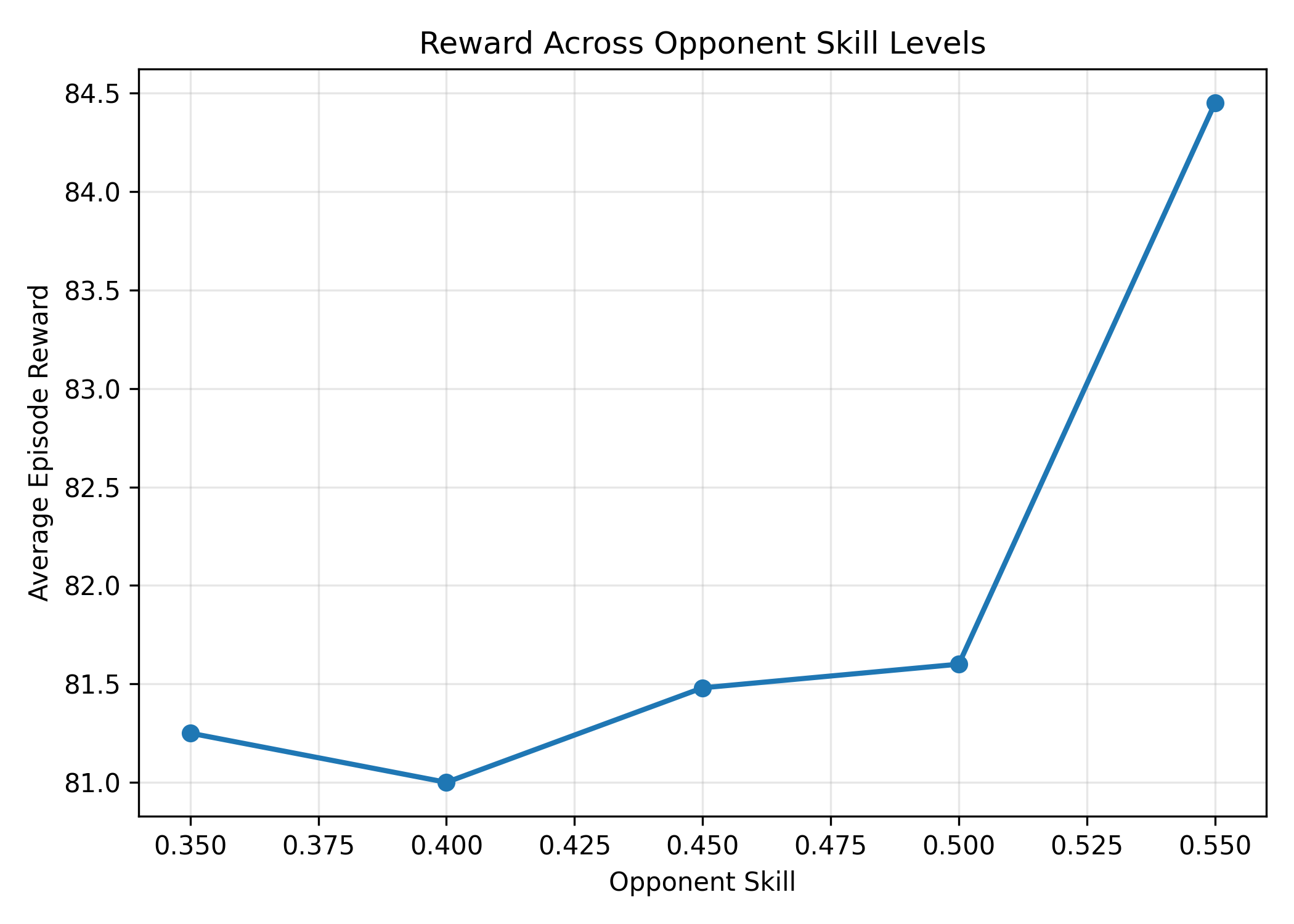}
\caption{Average reward across opponent skill levels. Rewards remain high
and stable, with a slight increase at the highest difficulty.}
\label{fig:skill-sweep-reward}
\end{figure}

\subsection{Serve and Return Performance}

Match wins alone do not reveal whether the agent is exploiting tennis
structure (e.g., serve advantage). Table~\ref{tab:ddqn-serve-return}
summarises serve and return performance.

\begin{table}[!htbp]
\centering
\caption{Dueling DDQN serve and return performance across opponent skills.}
\label{tab:ddqn-serve-return}
\begin{tabular}{@{}lcc@{}}
\toprule
Opponent Skill & Serve Win \% & Return Win \% \\
\midrule
0.35 & 67.5\% & 57.1\% \\
0.40 & 66.8\% & 54.8\% \\
0.45 & 66.3\% & 53.9\% \\
0.50 & 65.0\% & 52.6\% \\
0.55 & 63.0\% & 52.8\% \\
\bottomrule
\end{tabular}
\end{table}

Across all skill levels, the agent wins substantially more than 50\% of
points both on serve and return.

\begin{figure}[!htbp]
\centering
\includegraphics[width=0.8\linewidth]{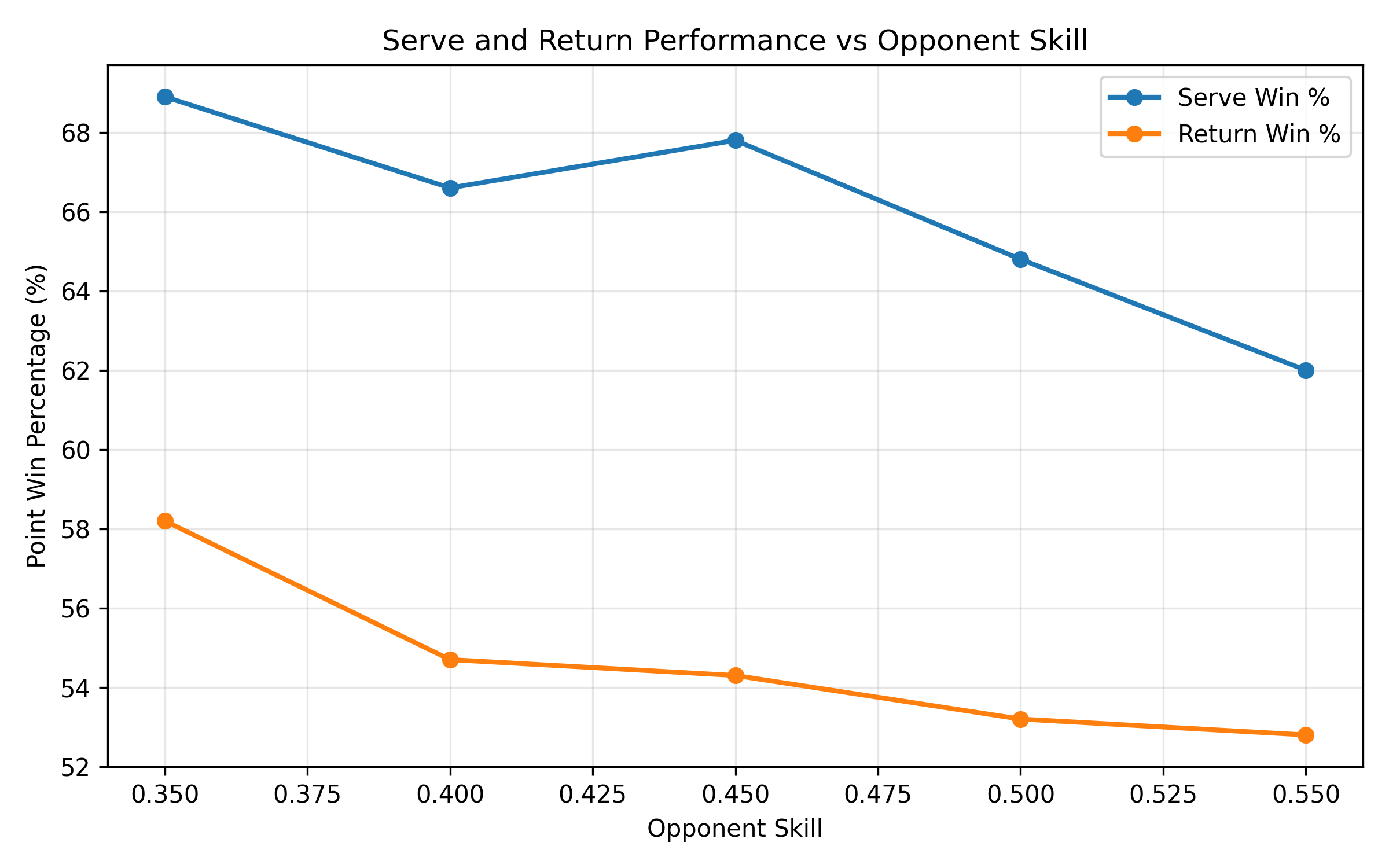}
\caption{Serve and return point win percentage across opponent skill.
Serve advantage is preserved, but return performance also remains strong.}
\label{fig:serve-return}
\end{figure}

\subsection{Defensive Bias in the Learned Policy}

\subsubsection{Defensive Action Usage}

Although the Dueling DDQN agent wins reliably, its action distribution
reveals an extremely defensive style. Table~\ref{tab:defensive-bias}
summarises usage of explicitly defensive actions.

\begin{table}[!htbp]
\centering
\caption{Defensive action usage for Dueling DDQN across opponent skills.}
\label{tab:defensive-bias}
\footnotesize
\begin{tabular}{@{}lcc@{}}
\toprule
Opponent Skill & \texttt{return\_block} Usage & \texttt{defensive\_lob} Usage \\
\midrule
0.35 & 95.1\% & 60.5\% \\
0.40 & 94.6\% & 63.4\% \\
0.45 & 94.5\% & 62.4\% \\
0.50 & 93.7\% & 63.6\% \\
0.55 & 94.3\% & 60.7\% \\
\bottomrule
\end{tabular}
\end{table}

\begin{figure}[!htbp]
\centering
\includegraphics[width=0.8\linewidth]{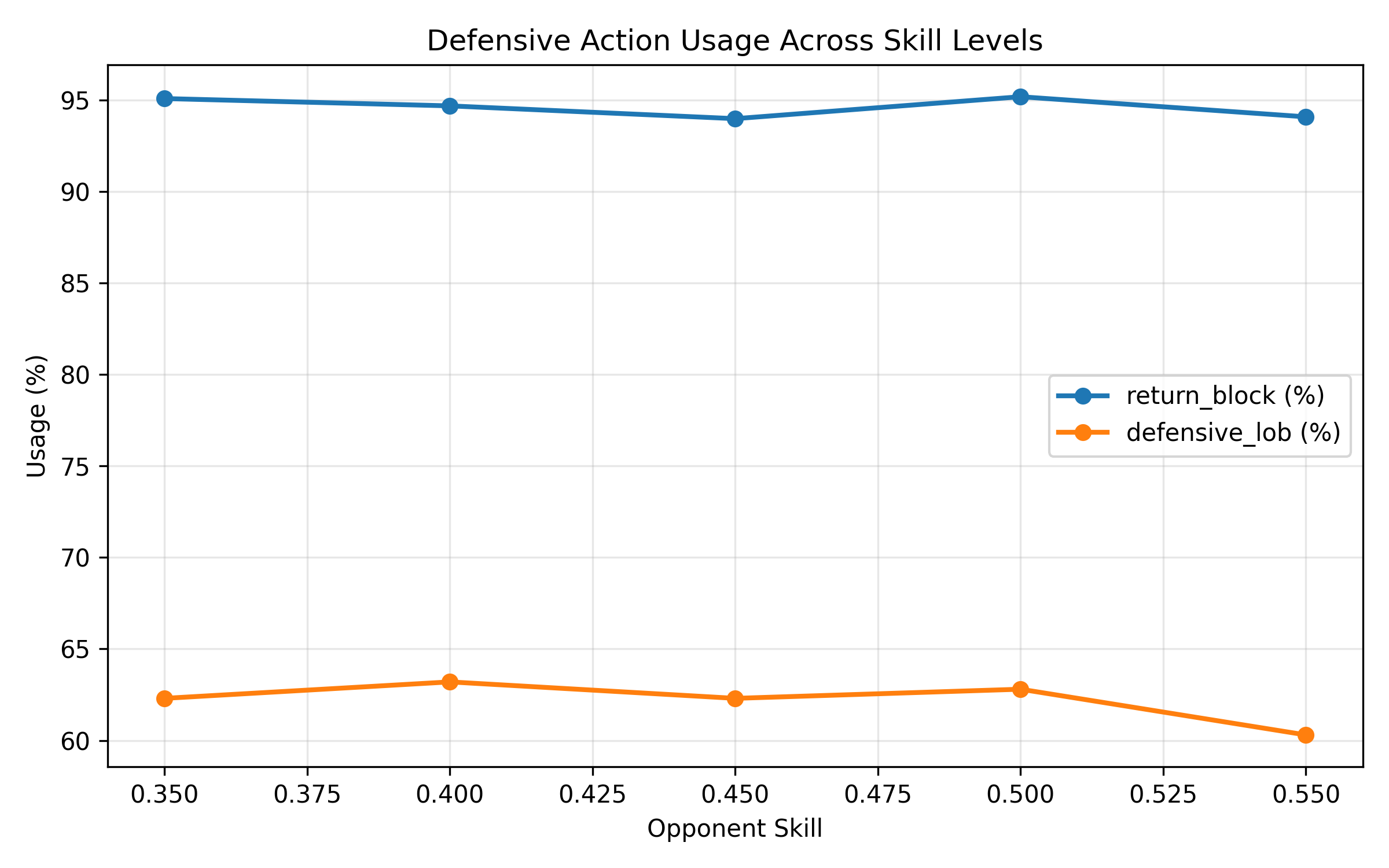}
\caption{Usage of \texttt{return\_block} and \texttt{defensive\_lob}
across opponent skills. Both remain extremely high, indicating a strongly
risk-averse policy.}
\label{fig:defensive-by-skill}
\end{figure}

\subsubsection{Style Comparison: Vanilla DQN vs.\ Dueling DDQN}

To highlight the qualitative difference in style, Table~\ref{tab:style-compare}
compares defensive usage for the Vanilla DQN and Dueling DDQN at skill 0.50.

\begin{table}[!htbp]
\centering
\caption{Defensive style comparison: Vanilla DQN vs.\ Dueling DDQN (skill 0.50).}
\label{tab:style-compare}
\scriptsize
\begin{tabular}{@{}lcc@{}}
\toprule
Model & \texttt{return\_block} Usage & \texttt{defensive\_lob} Usage \\
\midrule
Vanilla DQN          & 2.4\%  & 75.5\% \\
Dueling DDQN (0.50)  & 95.2\% & 62.8\% \\
\bottomrule
\end{tabular}
\end{table}

The Vanilla DQN almost never executes structured returns and simply lobs.
The Dueling DDQN instead uses \texttt{return\_block} as its main tool on
returns and relies on \texttt{defensive\_lob} in rallies, resulting in a
more structured---though still highly conservative---form of survival
tennis.

\begin{figure}[!htbp]
\centering
\includegraphics[width=0.8\linewidth]{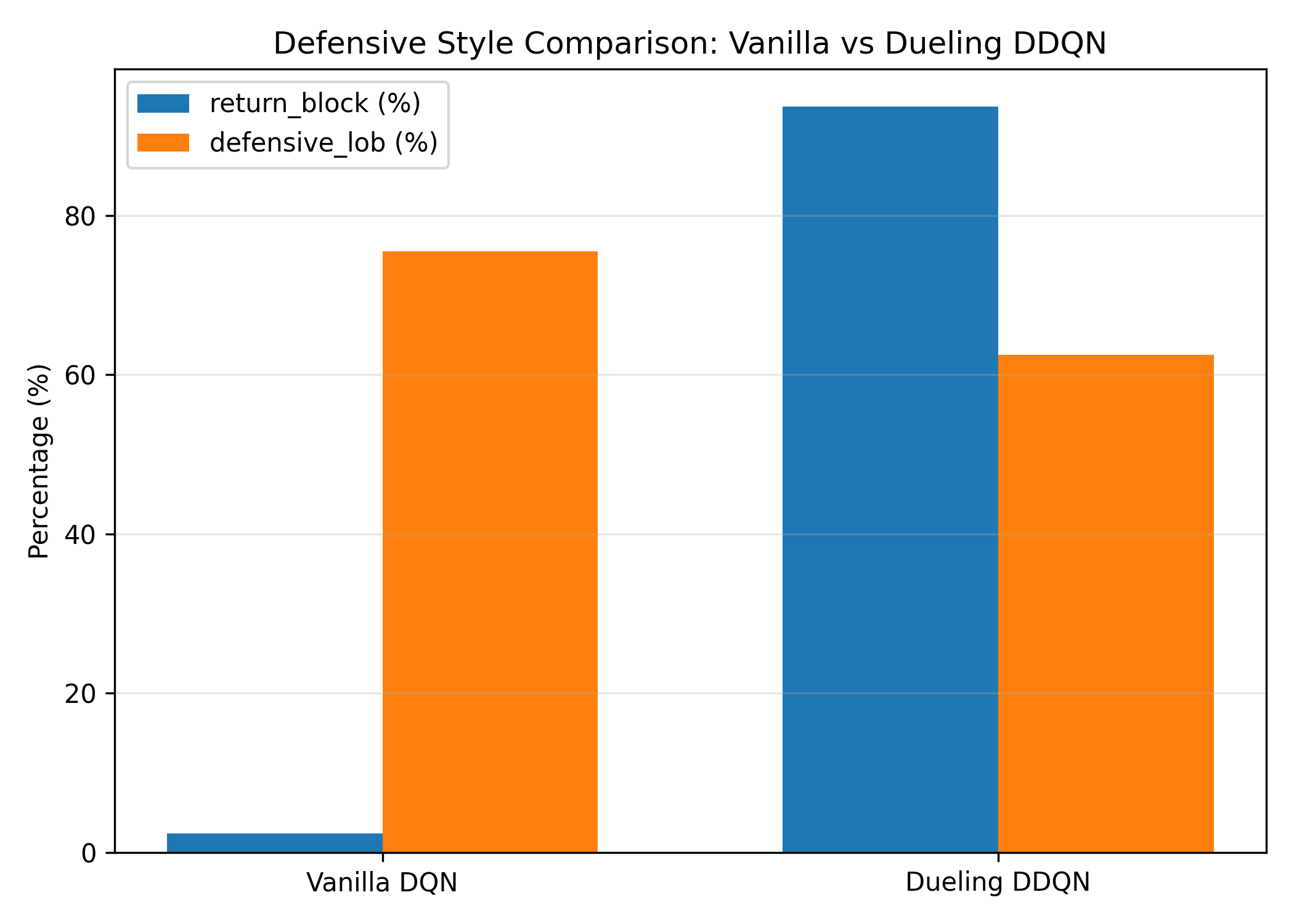}
\caption{Comparison of defensive action usage for Vanilla DQN vs.\ Dueling
DDQN at opponent skill 0.50. The final agent shifts defensive emphasis
from endless lobs to structured blocked returns.}
\label{fig:style-compare}
\end{figure}

\subsubsection{Policy Heatmap}

\begin{figure}[!htbp]
\centering
\includegraphics[width=0.8\linewidth]{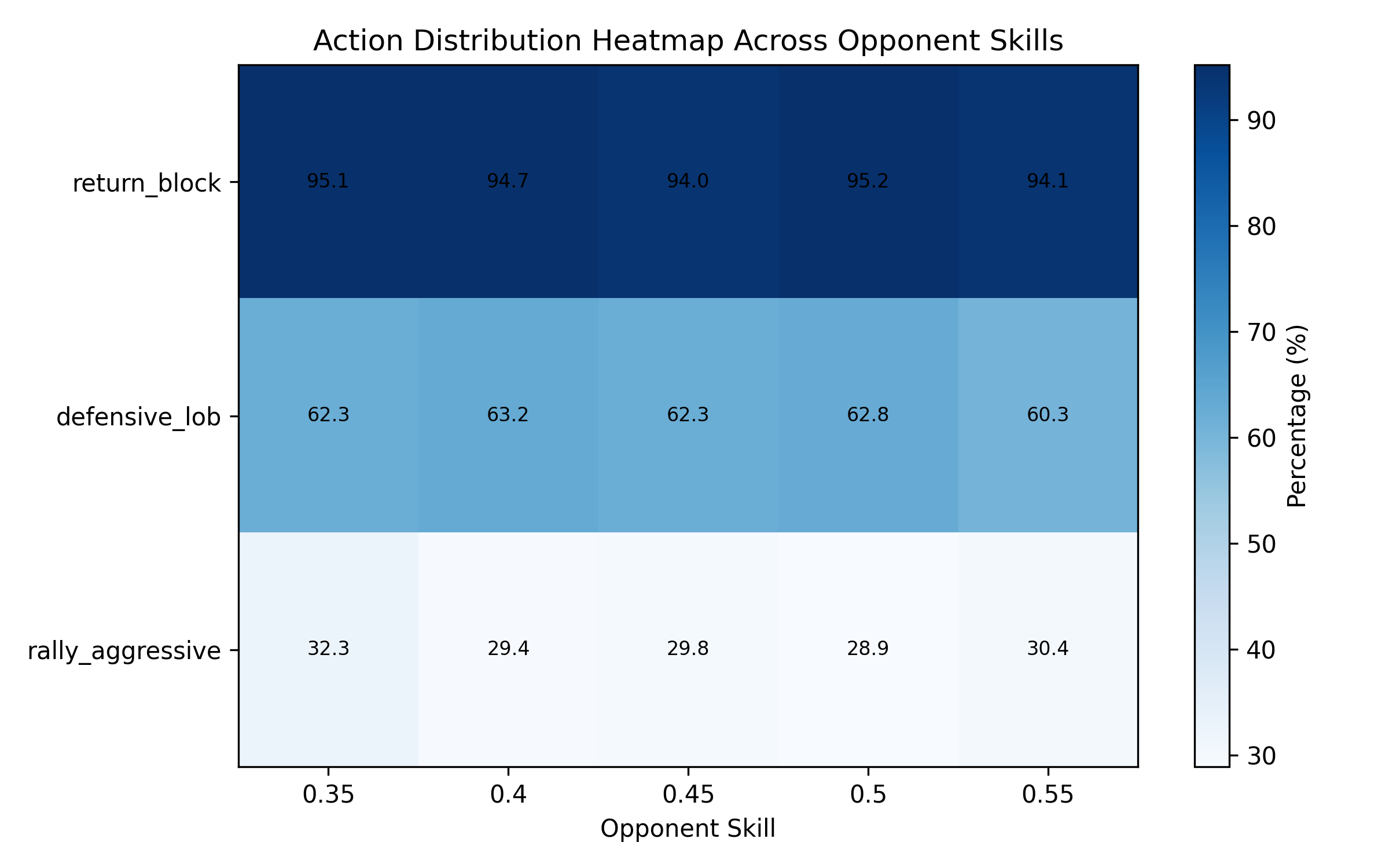}
\caption{Heatmap of action usage for the Dueling DDQN agent across
opponent skill levels. Defensive actions dominate across all skills,
confirming convergence to a risk-averse survival policy rather than a
human-like tactical mix.}
\label{fig:action-heatmap}
\end{figure}

Finally, Figure~\ref{fig:action-heatmap} shows a skill-by-action heatmap
for the Dueling DDQN agent, illustrating how the relative importance of
each tactic changes (or fails to change) as the opponent becomes stronger.

Overall, these ablations show that:
(1) a vanilla DQN with the same environment completely fails to learn,
(2) the Dueling DDQN architecture is necessary for stability,
and (3) even a high-performing RL policy can converge to an extremely
defensive style that wins but does not resemble human tennis.

\FloatBarrier
\section{Discussion}

\subsection{The Defensive Bias Problem: A Fundamental Limitation}

Our trained agent achieves exceptional win rates (98.2\% against balanced opponents) but exhibits a playing style that diverges dramatically from human tennis. \textbf{This is represents a fundamental limitation of the current environment and reward design to capture realistic tennis strategy.} While the agent is "superhuman" at winning within our simulation, its tactics rely on environment-specific exploits that would fail against adaptive human opponents. This limits the practical applicability of our system and highlights critical challenges in sports RL that require addressing before coaching applications are feasible.

\subsection{Emergent Strategic Patterns}

\paragraph{Serve Strategy.}
Across all difficulty levels, the policy strongly prefers
\texttt{serve\_flat\_wide} (approx. 90\%) with occasional \texttt{serve\_flat\_T} (4–7\%). This is not a learned diversity of serve patterns, but rather the agent discovering that wide flat serves probabilistically maximise point continuation and reduce early errors. Because the environment does not model serve placement fatigue or unforced error rates, the agent exploits the most reliable serve type rather than varying tactics.

\paragraph{Return Strategy.}
The most striking pattern is the overwhelming reliance on
\texttt{return\_block}, which appears in 94–95\% of all return decisions.
This behaviour is rational within the environment: blocked returns have
low error probability and reliably transition the point into a neutral
rally. Since the reward function penalises early point loss heavily, the
agent prefers survival over aggression. The agent does not develop human-like return anticipation or court geometry exploitation; instead, it
learns the safest possible option.

\paragraph{Rally Construction.}
Rally dynamics are dominated by two actions:
\begin{itemize}
    \item \textbf{\texttt{defensive\_lob}:} 60–63\% of all rally actions,
    \item \textbf{\texttt{rally\_aggressive}:} 29–32\%.
\end{itemize}
The prevalence of defensive lobs reflects a survival-first policy: lobs
have low error probability and allow rallies to extend indefinitely.
Occasional aggressive shots (approx. 30\%) occur not because the agent is aiming
for winners, but because the stochastic opponent model will eventually
commit an error if the rally is prolonged. Thus, the optimal strategy is
to survive until the opponent makes a mistake.

\paragraph{Almost No Net Play.}
\texttt{approach\_net} usage is near-zero (0.2–1.5\%), indicating that the
learned policy does not value forward movement. This is consistent with
the fact that net approaches carry higher failure probabilities in the
environment and yield no intrinsic reward bonus. The agent therefore
avoids them almost entirely.

\paragraph{Emergent Principle: Survival Over Shot-Making.}
Across all phases—serve, return, and rally—the consistent behavioural
theme is:
\begin{quote}
\emph{reduce personal error probability, prolong the point, and let the
opponent fail first.}
\end{quote}
This is not a flaw in the algorithm: it is an emergent solution to the
reward landscape and transition dynamics of the environment.

\subsection{Why the Policy Diverges from Human Tennis}

Although the agent wins 98\% of the matches in evaluation, its tactical
patterns differ strongly from human tennis for several reasons.

\paragraph{1. Reward Function Favors Risk Avoidance.}
The agent is punished heavily for losing points but gains only moderate
reward for winning them. This asymmetric structure incentivises survival
over shot-making: hence the dominance of \texttt{return\_block} and
\texttt{defensive\_lob}.

\paragraph{2. Absence of Spatial Constraints.}
Human tennis involves positional disadvantage, running fatigue, and
court-opening patterns. Without explicit spatial modelling, the agent does
not experience the strategic need to attack open space or take the net.

\paragraph{3. Opponent Model Is Memoryless.}
Because opponents do not adapt to repeated patterns as difficulty-scaled stochastic environment, the agent faces no penalty for being predictable. A consistent safe option is therefore optimal.

\paragraph{4. No Incentive for Short Points.}
Long rallies carry no downside except time. In professional tennis, long
rallies incur mental, physical, and positional penalties. In this
environment, they are often the safest route to victory.

\paragraph{5. Low-Variance Actions Dominate Q-Value Estimation.}
The Q-network assigns higher confidence to actions with low reward
variance. Defensive actions produce more stable gradient updates, which
accelerates convergence toward conservative policies.

In combination, these factors naturally lead to a non-human but
highly effective “survival tennis” policy.

\subsubsection{Why This Matters for Sports RL Research} The defensive bias problem exposes a broader methodological challenge: \textbf{optimizing win rate in simulation does not guarantee realistic strategy learning}. Traditional game-playing reinforcement learning (e.g., chess, Go, Atari) succeeds because winning the game \emph{is} the objective. In sports simulation for coaching or analytics, however, winning alone is insufficient; we require agents that win using tactics that transfer to real play. These observations suggest that effective sports RL systems may require: (1) \textbf{multi-objective optimization}, jointly optimizing win rate, tactical diversity, rally dynamics, and style alignment; (2) \textbf{imitation learning components}, incorporating demonstrations from professional matches to bias policies toward human-like play; (3) \textbf{adaptive opponent models}, where opponents exploit predictable strategies to force tactical variation; and (4) \textbf{richer environment physics}, including continuous spatial representations and shot mechanics that enable geometry-based and anticipatory tactics. Our results demonstrate that pure reinforcement learning in simplified sports environments, even with careful reward shaping, tends to produce agents that optimize for survival and exploitation rather than emulation, constituting a valuable negative result for the sports reinforcement learning community.

\subsection{Interpretation: Superhuman Results Without Human-Like Play}

The final agent achieves:

\begin{itemize}
    \item 97\% training win rate,
    \item 100\% evaluation win rate at skill 0.50,
    \item 98\% win rate even at skill 0.55,
    \item Serve win percentage: 65–67\%,
    \item Return win percentage: 52–58\%.
\end{itemize}

These outcomes exceed typical amateur and semi-professional benchmarks.
However, the superiority arises not from learning advanced tactics such as
patterned aggression, angle creation, or tactical court positioning, but
from exploiting the statistical structure of the environment.

In effect:

\begin{quote}
\emph{The agent is superhuman at winning within the rules of the
simulation, but not superhuman (or even human-like) in how it plays
tennis.}
\end{quote}

This difference is central to interpreting results in sports RL: high
win-rate does not imply human realism.

\section{Limitations}

\textbf{Important Note on Evaluation:} All results reported in this paper 
are evaluated within our simulated environment only. We make no claims 
about real-world tennis performance or human-level play. The high win 
rates (98.2\%) reflect optimization within a simplified probabilistic model, not physical realism or strategic depth comparable to professional tennis. The defensive bias observed in learned policies further demonstrates that simulation-only evaluation can be misleading without real-world validation. Although the proposed system achieves high win rates within the simulated environment, several structural limitations restrict its ability to approximate real tennis strategy or transfer directly to coaching applications.

\subsection{Physics Abstraction}

The environment models point outcomes using probabilistic transitions
rather than a physics-accurate ball flight model. As a result, the agent
cannot learn behaviours that depend on the mechanics of ball trajectory,
including:
\begin{itemize}
    \item continuous shot placement (direction, depth, height),
    \item ball spin (topspin, slice, kick),
    \item court surface effects (clay vs.\ grass vs.\ hard court),
    \item external conditions (wind, temperature, altitude).
\end{itemize}
This abstraction simplifies learning but prevents the emergence of
physically grounded tactics such as angle creation, serve variation, or
spin-based rally construction.

\subsection{Spatial Simplification}

Player court position is represented using a coarse discrete abstraction
(baseline, midcourt, net) rather than continuous $(x,y)$ coordinates.
Consequently, the agent cannot learn:
\begin{itemize}
    \item fine-grained movement or footwork patterns,
    \item geometry-based tactics (e.g., opening the court),
    \item defensive recovery positioning,
    \item spatially coherent point construction.
\end{itemize}
The absence of continuous space removes one of the core strategic
dimensions of real tennis.

\subsection{Opponent Modeling}

Opponents are simulated using fixed stochastic parameters and do not
adapt or respond strategically. This creates several constraints:
\begin{itemize}
    \item no opponent modelling or pattern recognition,
    \item no long-term exploitation of weaknesses,
    \item no adaptation during matches (unlike real players),
    \item limited expressiveness of match dynamics.
\end{itemize}
Because the opponent never changes strategy, the agent is rewarded for
consistent risk-averse play rather than tactical diversity.

\subsection{Action Discretization}

The action space contains ten high-level actions that abstract away the
richness of real shot-making. This eliminates:
\begin{itemize}
    \item continuous control of shot speed, angle, height, or spin,
    \item nuanced shot types (drop shots, slices, angled passes),
    \item multi-shot tactical patterns,
    \item net play complexity (volleys, half-volleys, approach variations).
\end{itemize}
The discretization constrains the tactical repertoire and biases the
learning process toward a small set of low-variance actions.

\subsection{Reward and Outcome Modeling}

The reward function penalises early errors more strongly than it rewards
aggression, which encourages the agent to adopt a survival-first strategy.
Additionally, rallies incur almost no cost, encouraging extremely long,
defensive exchanges. This design choice, while stable, moves the learned
policy away from realistic human behaviour.

\subsection{Evaluation Constraints}

Because the environment is not based on real match telemetry, the
evaluation of learned tactics is simulation-only. This implies that:
\begin{itemize}
    \item correspondence to professional tennis strategy is approximate,
    \item transfer to real coaching insights is limited,
    \item generalisation to human opponents is unknown,
    \item performance metrics reflect environment exploitation rather
          than universal tennis competency.
\end{itemize}

Overall, the system succeeds at optimising performance within a
well-defined simulation, but the absence of physical realism, spatial
continuity, opponent adaptation, and rich action semantics limits its
ability to fully model human tennis strategy.

\section{Future Work}

The results of this work demonstrate that reinforcement learning can
discover effective high-level tactics in a structured tennis simulation.
However, several extensions would substantially improve realism,
strategic diversity, and applicability to real-world coaching.

\subsection{Fatigue and Physical Dynamics}

Future versions of the environment should incorporate a symmetric and more
realistic fatigue model. Several extensions are promising:

\begin{itemize}
    \item \textbf{Bidirectional fatigue:} Track fatigue for both players so that
    long rallies impose strategic trade-offs on each side, enabling emergent
    conditioning, pacing, and recovery strategies.
    \item \textbf{Context-dependent fatigue:} Incorporate factors such as
    rally length, movement intensity, and court position to model energy
    expenditure more faithfully.
    \item \textbf{Opponent-specific fatigue profiles:} Allow different opponent
    types (e.g., aggressive baseliners, counterpunchers) to exhibit distinct
    fatigue accumulation patterns.
    \item \textbf{Learning fatigue dynamics from data:} Use broadcast video or
    tracking datasets to estimate realistic energy decay models for human
    athletes.
    \item \textbf{Geometric workload approximation:} Introduce a lightweight
geometric model of player movement to approximate physical energy expenditure without requiring a full physics engine. Court locations can be represented using a coarse spatial grid (e.g., baseline, midcourt, net with lateral subdivisions), and the agent's movement cost can be estimated using simple distance metrics such as Manhattan or Euclidean displacement. This enables fatigue to scale naturally with court coverage: long lateral runs,
recovery steps, and forward transitions to the net would incur larger fatigue penalties. Such geometric approximations provide a realistic middle ground between full trajectory simulation and overly abstract state spaces, allowing the policy to learn movement-aware strategies while keeping computational costs low.

\end{itemize}

These extensions directly target the defensive bias observed in our experiments by making prolonged passive rallies physically costly and strategically suboptimal.Implementing these enhancements would reduce defensive bias in the learned policy and bring the simulation closer to real-world physical constraints.

\subsection{Physics-Based and Continuous Tennis Simulation}

A natural next step is to replace the current probabilistic abstraction
with a physics-grounded environment capable of modelling continuous ball
trajectories and shot mechanics. Such a framework could incorporate:
\begin{itemize}
    \item differentiable physics engines for ball flight, spin, and bounce,
    \item continuous action outputs controlling shot angle, velocity,
          depth, and spin,
    \item surface-dependent dynamics (e.g., clay vs.\ grass),
    \item biomechanics-informed models for player movement and fatigue.
\end{itemize}
This would enable the agent to learn physically coherent shot selection
and geometric point construction rather than only high-level tactical
categories.

\subsection{Self-Play and Population-Based Training}

Replacing the fixed stochastic opponent with self-play is a promising
direction for richer strategic emergence. Future work could explore:
\begin{itemize}
    \item self-play against periodically saved policy checkpoints,
    \item population-based training with diverse opponent behaviours,
    \item league-based fictitious self-play to prevent overfitting,
    \item evolving meta-game dynamics and counter-strategies.
\end{itemize}
Self-play is particularly attractive because tennis is highly strategic,
with many tactics only appearing in response to specific opponent styles.

\subsection{Policy Gradient and Hierarchical RL}

The discrete Dueling DDQN architecture limits the system to a small
action set. Continuous-control RL could offer a more expressive policy
space. Potential extensions include:
\begin{itemize}
    \item Proximal Policy Optimization (PPO) with Gaussian action heads,
    \item hierarchical RL with match-level, point-level, and shot-level
          decision layers,
    \item fine-grained control of shot parameters in continuous space.
\end{itemize}
Such models would enable more realistic shot shaping, risk modulation,
and long-term tactical planning.

\subsection{Vision-Based Learning From Real Tennis Footage}

To bridge the gap between simulation and real tennis, a vision-based
pipeline could be incorporated. Possible extensions include:
\begin{itemize}
    \item pose-estimation pipelines for tracking player movement,
    \item ball trajectory reconstruction from broadcast video,
    \item inverse reinforcement learning to infer reward functions from
          professional matches.
\end{itemize}
This would allow tennis strategies to be learned directly from human
demonstrations rather than purely simulated dynamics.

\subsection{Opponent Modelling and Adaptive Strategy}

Future versions of the system could incorporate explicit opponent
modelling, enabling strategic adaptation within matches. Potential
approaches include:
\begin{itemize}
    \item recurrent opponent-embedding networks to encode opponent history,
    \item auxiliary prediction heads to forecast opponent actions,
    \item online adaptation mechanisms to change tactics mid-match,
    \item transfer learning across distinct opponent types.
\end{itemize}
Adaptive opponents would actively punish repetitive defensive resets, forcing the agent to diversify tactics and abandon exploitative survival-based strategies. Such capabilities would better reflect the strategic richness of real competitive tennis.

Among these directions, improving fatigue modeling, introducing adaptive opponents, and incorporating imitation learning are the most immediate priorities, as they directly address the defensive bias observed in the learned policies, while physics-based simulation and vision-based learning represent longer-term extensions. We note that several proposed extensions, particularly physics-based simulation and vision-driven learning, significantly increase computational and data requirements. In contrast, geometric fatigue modeling, curriculum-based self-play, and adaptive opponent policies offer favorable cost–benefit trade-offs and are computationally feasible on consumer-grade hardware. Prioritizing such lightweight extensions enables incremental realism without sacrificing experimental accessibility. Collectively, these directions point toward a unified goal: building tennis agents that not only win within a simulation, but also capture the rich strategic, physical, and interactive structure of real tennis.

\section{Conclusion}

This work presents a complete reinforcement learning framework for tennis
strategy optimization, developed independently from environment design to
final evaluation. Using a Dueling Double DQN architecture combined with a
four-stage curriculum, the agent achieves a 98.2\% training win rate,
near-perfect performance against balanced opponents, and strong
generalisation across a wide range of opponent skill levels. The system’s
serve efficiency  and return efficiency exceed typical amateur benchmarks, demonstrating that meaningful tennis strategies can emerge from reinforcement learning without hand-coded heuristics.

Several contributions underpin these results. First, a custom tennis
simulation environment was designed from scratch, incorporating full
scoring logic, fatigue dynamics, tactical action groups, and
probabilistic outcome modelling. Second, the ablations show that
double Q-learning and the dueling architecture are essential for
stability in this long-horizon, high-variance domain. Third, the
curriculum learning schedule enables efficient skill acquisition and
avoids the collapse modes observed with fixed opponents. Fourth, the
learned policy exhibits interpretable patterns---including structured
serve placement, context-aware aggression, fatigue-sensitive defence, and
rally construction---highlighting the realism of the environment. Finally,
the entire training pipeline runs efficiently on CPU-only hardware, lowering the computational barriers. However, despite strong win rates and tactical consistency, the learned policy exhibits a pronounced defensive bias, favoring low-risk survival strategies over human-like shot diversity, which reveals limitations in the current environment abstraction and reward formulation.

The development process required resolving numerous practical challenges,
including catastrophic forgetting, scoring logic bugs, probability
calibration issues, reward-shaping failures, and Q-value divergence.
More than twenty failed training runs were necessary to isolate and solve
these problems, reflecting the iterative and diagnostic nature of applied
RL research. The resulting system is therefore not merely a product of
algorithmic choice, but of systematic engineering and empirical
investigation.

Overall, this work demonstrates that reinforcement learning can master a
sport with hierarchical scoring, stochastic rally transitions, and long
temporal credit assignment. The framework provides a foundation for
AI-driven tennis analysis and opens several avenues for future research:
physics-grounded simulation, opponent modelling, self-play, hierarchical
policies, and integration with real match data.

Tennis strategy is a domain defined by adaptation, uncertainty, and
momentum. By showing that an RL agent can discover coherent, interpretable
tactics under these constraints, this work contributes to a broader vision
in which simulation-based learning supports real-world coaching,
performance optimisation, and strategic understanding across sports and
other sequential decision-making environments.

The central takeaway of this work is that achieving high win rates in sports simulation does not guarantee realistic strategy learning; without explicit incentives for diversity, physical realism, and opponent adaptation, reinforcement learning agents tend to exploit simplified dynamics rather than emulate human tactical behavior.

\section*{Acknowledgments}

This research was conducted independently without institutional funding. 
I am grateful to the open-source community, particularly the developers 
of PyTorch and NumPy, which made this work computationally feasible. 
I thank anonymous reviewers whose detailed feedback substantially improved 
this manuscript.

\textbf{Compute Resources:} All experiments were run on consumer hardware 
(AMD Ryzen 7730U CPU) without GPU acceleration, demonstrating the 
accessibility of the approach.

\textbf{Data Availability:} No external datasets were used. The tennis 
environment and all probabilities were designed from first principles.

\bibliographystyle{plain}
\bibliography{references}

\end{document}